\documentclass[a4paper,fleqn]{cas-dc}

\usepackage[numbers]{natbib}

\usepackage{amssymb}
\usepackage{amsmath}

\usepackage{hyperref}
\usepackage{graphicx}

\usepackage{subcaption}
\usepackage{doi}
\usepackage[switch]{lineno}
\usepackage[textwidth=3.5cm]{todonotes}

\usepackage{booktabs}
\usepackage{multirow}
\usepackage{xcolor}
\usepackage{colortbl}
\usepackage{pgf}

\usepackage{tikz}
\usetikzlibrary{arrows.meta}

\def\tsc#1{\csdef{#1}{\textsc{\lowercase{#1}}\xspace}}
\tsc{WGM}
\tsc{QE}

\begin{document}
\let\WriteBookmarks\relax
\renewcommand{\topfraction}{0.9}
\renewcommand{\textfraction}{0.1}
\renewcommand{\floatpagefraction}{0.8}

\shorttitle{Flexible-body Modeling, Kinematic Identification, and Assembly Accuracy of Overconstrained Spatial Linkages}    

\shortauthors{Huczala, Pieber et al.}  

\title [mode = title]{Flexible-body Modeling, Kinematic Identification, and Assembly Accuracy of Overconstrained Spatial Linkages}  



\author[1,5]{Daniel Huczala}[orcid=0000-0002-7398-7825]
\affiliation[1]{organization={Department of Mechanical Engineering, Ulsan
National Institute of Science and Technology (UNIST)},
            addressline={50 UNIST-gil, Eonyang-eup, Ulju-gun},
            city={Ulsan},
            postcode={44919},
            country={South Korea}}
            
\credit{Methodology, Software, Data curation, Writing - Original draft preparation}

\author[2]{Michael Pieber}[orcid=0000-0002-3745-3439]

\credit{Laboratory Experiments, Methodology, Software, Data curation, Writing - Original draft preparation}

\author[2]{Johannes Gerstmayr}[orcid=0000-0003-3576-1830]
\affiliation[2]{organization={Department of Mechatronics, University of Innsbruck},
            addressline={Technikerstraße~13},
            city={Innsbruck},
            postcode={6020},
            country={Austria}}

\credit{Conceptualization of this study, Software}

\author[3]{Andreas Mair}[orcid=0009-0006-9955-1655]

\credit{Methodology, Software, Writing - Original draft preparation}

\author[3]{Frederik Schulte}[orcid=0009-0009-4594-3815]

\credit{Laboratory Experiments, Methodology, Writing - Original draft preparation}

\author[3]{Silvia Glas}
\affiliation[3]{organization={Unit of Geometry and Surveying, University of Innsbruck},
            addressline={Technikerstraße~13},
            city={Innsbruck},
            postcode={6020},
            country={Austria}}

\credit{Laboratory Experiments}

\author[4]{Tomas Postulka}[orcid=0009-0009-9453-9958]
\affiliation[4]{organization={Department of Robotics, VSB -- Technical University of Ostrava},
            addressline={17.~listopadu~2172/15},
            city={Ostrava},
            postcode={70800},
            country={Czech Republic}}

\credit{Methodology, Writing - Original draft preparation}

\author[5]{Ales Vysocky}[orcid=0000-0001-6942-4280]
\affiliation[5]{organization={Department of Machine Parts and Mechanisms, VSB -- Technical University of Ostrava},
            addressline={17.~listopadu~2172/15},
            city={Ostrava},
            postcode={70800},
            country={Czech Republic}}

\credit{Conceptualization of this study, Methodology, Writing - Original draft preparation}

\author[3]{Martin Pfurner}[orcid=0000-0003-1988-2202]
\cormark[1]
\ead{martin.pfurner@uibk.ac.at}

\credit{Conceptualization of this study, Methodology, Writing - Original draft preparation}

\cortext[1]{Corresponding author}



\begin{abstract}
Overconstrained rational single-loop linkages are efficient, compact, and low-cost custom mechanisms, yet their deployment in industrial settings is limited. In simulations, rigid body formulations fail due to redundant constraints. This study presents a flexible multibody modeling framework based on the floating frame of reference formulation, and delivers an overall accuracy analysis of assembled linkages prototypes. The approach is validated against 3D-printed PLA prototypes of a Bennett four-bar mechanism, including variants with intentional joint-axis misalignment, which theoretically, from the rigid body point of view, cannot be assembled. A supplementary contribution is delivered in the form of a kinematic parameter identification methodology suited for this type of mechanism with ill-conditioned Jacobian. The experimental and simulation results are compared and reveal that these overconstrained mechanisms exhibit a self-assembling tendency -- structural compliance drives the assembly toward the ideal geometric configuration, distributing constraint stress throughout the structure. Additional qualitative demonstrations using cardboard tubes and bamboo sticks as link building blocks confirm that functional mechanisms can be realized from low-cost, unconventional materials with limited manufacturing accuracy. The proposed modeling pipeline is fully algorithmic and enables design optimization in the future.
\end{abstract}





\maketitle

\newcommand{\qi}{\mathbf{i}}
\newcommand{\qj}{\mathbf{j}}
\newcommand{\qk}{\mathbf{k}}

\section{Introduction}


Classic six-revolute joint (6R) serial robots are widely used in automation as universal manipulators capable of handling a variety of tasks. However, this universality comes at a cost: higher energy consumption, greater expense, complex control, and the need for regular maintenance. For companies seeking to automate a specific task within a limited budget, \emph{custom mechanisms} can be a more suitable solution. Although single-purpose by design, these task-specific mechanisms can outperform standard robots in areas including, but not limited to, energy efficiency, speed, and accuracy. It is a widely studied research direction~\cite{Bansal2021review} and this paper focuses on one such type of custom mechanism: overconstrained rational single-loop linkages with 1 Degree of Freedom (DoF). A well-known example is the Bennett four-bar linkage~\cite{bennett1914skew}, which is the simplest spatial overconstrained mechanism parameterized by a rational motion.

\begin{figure}[pos=tb]
    \centering
    \includegraphics[width=\linewidth]{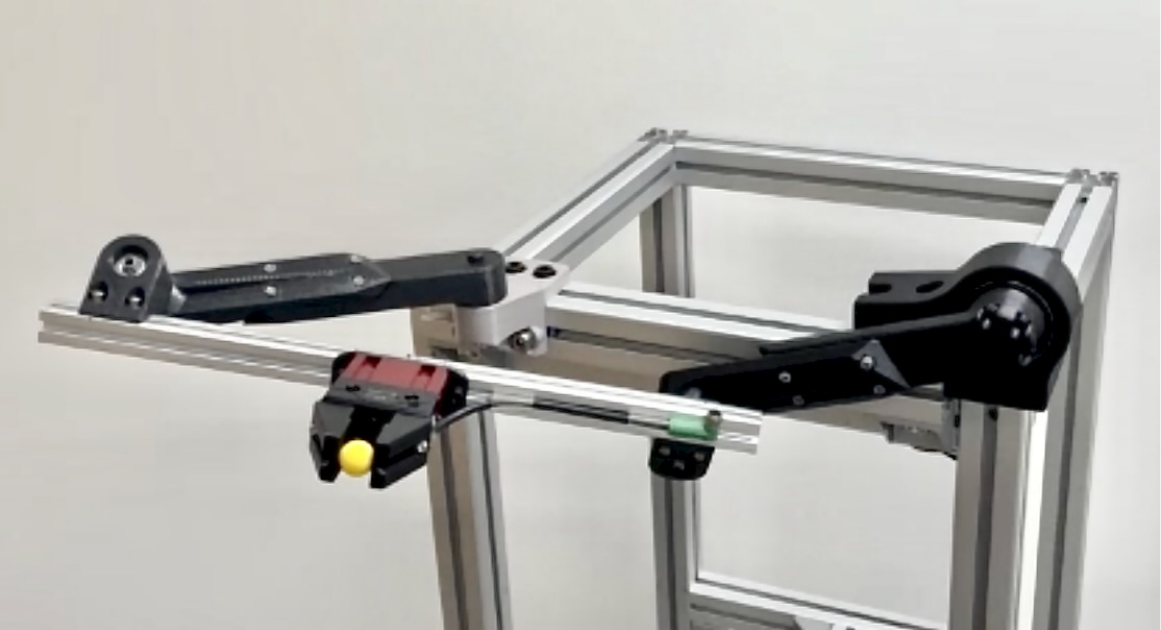}
    \caption{Laboratory manipulator prototype of a Bennett mechanism.}
    \label{fig:bennet_lab}
\end{figure}

The commercial deployment of the discussed mechanisms has so far been limited to laboratory mixers such as Turbula or Oloid, and robot manipulation devices are rather limited to research experiments, as shown in Fig.~\ref{fig:bennet_lab}. The reason was missing general methodology due to the specific design; however, this has been partially addressed in recent years: With the introduction of the rational motion factorization method~\cite{Hegeds2013} the task-based synthesis of these mechanisms became straightforward: 3 poses~\cite{brunnthaler2005new} or 5 points~\cite{Zube2018} in space yield a quadratic motion, i.e. four-bar mechanism; 4 poses~\cite{Hegeds2015} or 7 points yield a cubic motion~\cite{huczala2026-ark}, i.e. a six-bar mechanism also with one DoF. Self-intersection-free design was proposed in~\cite{LiNawratil2020}, and the kinematic motion planning and numerical inverse kinematics was reported in~\cite{huczala2024velocity}. In addition, all the methodology \emph{is implemented and available} as an open source Python package Rational Linkages~\cite{huczala2024linkages}. Note that all these geometrical and kinematic methods benefit from the genuine \emph{rational} representation of the motion, although they often utilize the numerical approach to obtain results faster. This study aims to deliver one of the last puzzles missing in the current state-of-the-art: a general multibody model. 

Single-loop linkages with one DoF can reduce running costs by saving up to 92\% in energy consumption~\cite{postulka2025effectiveness} compared to standard 6R robots while manipulating the same payload. Additionally, manufacturing costs can be significantly lower, particularly when alternative materials such as plastic, bamboo, or cardboard are used where the application permits. These mechanisms are lightweight and easy to transport, as they can be folded into compact configurations, making them suitable for a variety of applications beyond industrial automation -- for example, as movable sculptures or artificial trees that track the sun providing desired shade throughout the day~\cite{LiNawratil2020}, or foldable flapping bird wings in bio-inspored robotics~\cite{Huczala2026birdins}. They can also be coupled and integrated into deployable networks, to expand in desired surfaces~\cite{Lyu2022approxi} or for architectural design using bamboo as construction material in pavilion structures~\cite{SuzukiPauly2023}.

For many applications, particularly those involving high-speed motions, a purely kinematic approach is insufficient, as the control system must also account for moving masses and adjust the torques at the driving joint accordingly. However, applying the established standard multibody modeling method, rigid body modeling, is complicated in the case of overconstrained mechanisms~\cite{Mueller2009, Postulka2026analyti}, as the mobility criterion Chebychev–Grübler-Kutzbach yields $-2$ for four-bar mechanisms and $0$ for six-bar mechanisms, implying that they are, in general, impossible to assemble or are rigid, respectively.
The mobility is achieved only by additional geometrical conditions, such as the Bennett condition~\cite{bennett1914skew} for the four-bar mechanisms. The overconstrained property introduces redundant constraints into the rigid body system that are difficult to handle numerically. The simplest approach is to \emph{relax} some joints to obtain a non-overconstrained system, but the results immediately become questionable. There exist heuristic or semialgebraic ways to deal with this issue, such as constraints elimination~\cite{Mueller2014}, constraints decomposition~\cite{Yang2025acomput}, or introduce inequality constraints~\cite{Li2016dynamic} in the rigid body model. However, this requires performing kinematic analysis steps for a given type of mechanism prior to dynamics modeling, which can be suitable in control, but it is challenging for algorithmization and application in optimization strategies during the design phase.
Furthermore, as stated in studies dealing with redundant constraints~\cite{Frczek2011, GarcadeJaln2013}, a rigid body model of an overconstrained system does not correspond to real physics, and in order to find a unique set of all joint reaction forces, it is necessary to abandon the assumption that all bodies are rigid. Additionally, M{\"u}ller claims~\cite{Mueller2009} that in practice: ``Due to the flexibility of links and joint clearances, the real mechanism will exhibit almost the type of motion of its perfect (overconstrained) prototype... It is therefore important to allow for flexible members and joint clearances.'' 

In this work, we build upon the demonstration of flexible modeling capabilities for \emph{planar} four-bar mechanisms~\cite{Bauchau2016} and present a modeling framework for \emph{spatial} mechanisms that are beyond the planar case -- also overconstrained. Flexible bodies are modeled using the floating frame of reference formulation (FFRF)~\cite{Zwolfer2020, Zwoelfer2021}.
This study also validates the modeling approach against 3D-printed prototypes from polylactic acid (PLA) material, including variants with intentional misalignments. 
Using the latest available methodologies and tools, our pipeline from mechanism synthesis to flexible dynamics analysis is fully algorithmic, opening the door to future applications in optimal design.

Two supplementary contributions are introduced: a method for recovering rational motions from design parameters, and a novel kinematic parameter identification method for single-loop linkages, both used for validation purposes.
A systematic identification approach tailored to single-loop overconstrained mechanisms remains absent from the literature. The prevailing methodology for serial and parallel robots formulates calibration as a nonlinear optimization task, relying on external metrology systems (such as laser trackers) for high-accuracy end-effector pose measurements~\cite{Khanesar2023precisi, Wang2024methodf}. Parameter estimation employs techniques such as iterative least-squares, Singular Value Decomposition (SVD), and metaheuristic algorithms~\cite{Zhang2025kinemat}, minimizing the error between observed and predicted kinematics along a predefined trajectory via the robot Jacobian. This is not well suited to single-loop linkage architectures for two reasons. First, the overconstrained nature of the mechanism restricts the configuration space to a 1-DoF trajectory. Second, its geometric parameters are strictly coupled through the Bennett condition -- modifying any single parameter independently violates loop closure, yielding an ill-conditioned identification Jacobian. This paper proposes a methodology specifically adapted to these constraints, tracking all bodies of the mechanism and identifying the joint axes as lines in space using Principal Component Analysis (PCA) and circle fitting.

Last but not least, we investigate the hypothesis that even relatively inaccurate assemblies of these mechanisms can yield functional movable structures. To this end, prototypes made of cardboard tubes and bamboo sticks are included as qualitative demonstrations that overconstrained linkages can be constructed from unconventional, low-cost materials and still perform the desired motion, bringing the additional benefits of sustainability and low manufacturing cost. The insights gained from this experience are also applicable to other types of custom mechanisms.

The paper is structured as follows. Sect.~\ref{sec:methods} introduces the theoretical framework, including kinematics of overconstrained linkages, flexible multibody modeling via FFRF, and kinematic parameter identification. Sect.~\ref{sec:assembly_of_mechs} describes the prototypes (PLA, cardboard, bamboo). Sect.~\ref{sec:accuracy-analysis} details the simulation setup and experimental validation using motion capture. Sect.~\ref{sec:results} presents trajectory comparisons and DH parameter analysis. Finally, Sect.~\ref{sec:discussion} discusses the self-assembling behavior and future directions for six-bar mechanisms. In addition, supplementary material is published at~\cite{suppl_mat}, including data and corresponding analysis.

\section{Methods}\label{sec:methods}

The following subsections describe the theoretical framework that was used to carry out the proposed research.



\subsection{Kinematics of Overconstrained Single-loop Linkages}

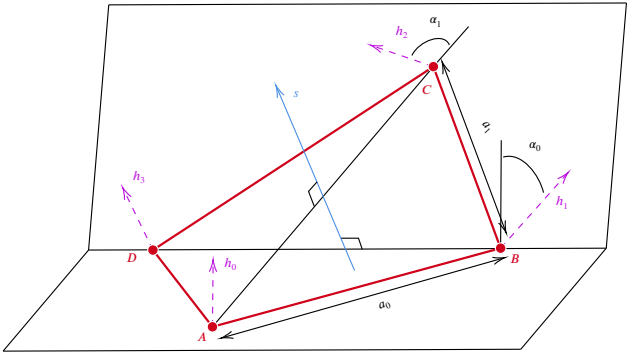
\begin{figure}[pos=tb]
    \centering
    \resizebox{\linewidth}{!}{\tikzset{every picture/.style={line width=0.75pt}} 

\begin{tikzpicture}[x=0.75pt,y=0.75pt,yscale=-1,xscale=1]

\draw    (104.8,241.13) -- (569.88,239.45) ;
\draw    (104.8,241.13) -- (28,332) ;
\draw    (569.88,239.45) -- (493.08,330.32) ;
\draw    (28,332) -- (493.08,330.32) ;
\draw    (122.92,20.68) -- (588,19) ;
\draw    (122.92,20.68) -- (104.8,241.13) ;
\draw    (588,19) -- (569.88,239.45) ;
\draw [color={rgb, 255:red, 208; green, 2; blue, 27 }  ,draw opacity=1 ][line width=1.5]    (162.61,241.13) -- (216.1,310.12) -- (474.96,239.45) ;
\draw [color={rgb, 255:red, 208; green, 2; blue, 27 }  ,draw opacity=1 ][line width=1.5]    (474.96,239.45) -- (414.56,76.22) -- (162.61,241.13) ;
\draw    (414.56,76.22) -- (216.1,310.12) ;
\draw [color={rgb, 255:red, 74; green, 144; blue, 226 }  ,draw opacity=1 ]   (343.81,259.64) -- (273.84,94.88) ;
\draw [shift={(273.05,93.04)}, rotate = 66.99] [color={rgb, 255:red, 74; green, 144; blue, 226 }  ,draw opacity=1 ][line width=0.75]    (10.93,-3.29) .. controls (6.95,-1.4) and (3.31,-0.3) .. (0,0) .. controls (3.31,0.3) and (6.95,1.4) .. (10.93,3.29)   ;
\draw    (331.73,230.47) -- (347.26,230.47) -- (350.71,240.57) ;
\draw    (310.16,179.71) -- (302.39,188.4) -- (307.57,201.86) ;
\draw [color={rgb, 255:red, 189; green, 16; blue, 224 }  ,draw opacity=1 ] [dash pattern={on 4.5pt off 4.5pt}]  (216.1,310.12) -- (216.68,251.68) ;
\draw [shift={(216.7,249.68)}, rotate = 90.56] [color={rgb, 255:red, 189; green, 16; blue, 224 }  ,draw opacity=1 ][line width=0.75]    (10.93,-3.29) .. controls (6.95,-1.4) and (3.31,-0.3) .. (0,0) .. controls (3.31,0.3) and (6.95,1.4) .. (10.93,3.29)   ;
\draw [color={rgb, 255:red, 189; green, 16; blue, 224 }  ,draw opacity=1 ] [dash pattern={on 4.5pt off 4.5pt}]  (474.96,239.45) -- (534.61,172.51) ;
\draw [shift={(535.94,171.01)}, rotate = 131.7] [color={rgb, 255:red, 189; green, 16; blue, 224 }  ,draw opacity=1 ][line width=0.75]    (10.93,-3.29) .. controls (6.95,-1.4) and (3.31,-0.3) .. (0,0) .. controls (3.31,0.3) and (6.95,1.4) .. (10.93,3.29)   ;
\draw [color={rgb, 255:red, 189; green, 16; blue, 224 }  ,draw opacity=1 ] [dash pattern={on 4.5pt off 4.5pt}]  (414.56,76.22) -- (359.85,57.5) ;
\draw [shift={(357.96,56.86)}, rotate = 18.88] [color={rgb, 255:red, 189; green, 16; blue, 224 }  ,draw opacity=1 ][line width=0.75]    (10.93,-3.29) .. controls (6.95,-1.4) and (3.31,-0.3) .. (0,0) .. controls (3.31,0.3) and (6.95,1.4) .. (10.93,3.29)   ;
\draw [color={rgb, 255:red, 189; green, 16; blue, 224 }  ,draw opacity=1 ] [dash pattern={on 4.5pt off 4.5pt}]  (162.61,241.13) -- (136.21,186.55) ;
\draw [shift={(135.34,184.75)}, rotate = 64.19] [color={rgb, 255:red, 189; green, 16; blue, 224 }  ,draw opacity=1 ][line width=0.75]    (10.93,-3.29) .. controls (6.95,-1.4) and (3.31,-0.3) .. (0,0) .. controls (3.31,0.3) and (6.95,1.4) .. (10.93,3.29)   ;
\draw [line width=0.75]    (225.52,319.81) -- (477.08,248.41) ;
\draw [shift={(479.01,247.86)}, rotate = 164.15] [color={rgb, 255:red, 0; green, 0; blue, 0 }  ][line width=0.75]    (10.93,-3.29) .. controls (6.95,-1.4) and (3.31,-0.3) .. (0,0) .. controls (3.31,0.3) and (6.95,1.4) .. (10.93,3.29)   ;
\draw [shift={(223.6,320.35)}, rotate = 344.15] [color={rgb, 255:red, 0; green, 0; blue, 0 }  ][line width=0.75]    (10.93,-3.29) .. controls (6.95,-1.4) and (3.31,-0.3) .. (0,0) .. controls (3.31,0.3) and (6.95,1.4) .. (10.93,3.29)   ;
\draw    (477.27,157.55) .. controls (494.52,152.5) and (510.05,174.38) .. (513.51,189.52) ;
\draw    (475.54,142.41) -- (474.96,239.45) ;
\draw  [color={rgb, 255:red, 255; green, 255; blue, 255 }  ,draw opacity=1 ][fill={rgb, 255:red, 208; green, 2; blue, 27 }  ,fill opacity=1 ] (157.87,241.13) .. controls (157.87,238.51) and (159.99,236.4) .. (162.61,236.4) .. controls (165.22,236.4) and (167.34,238.51) .. (167.34,241.13) .. controls (167.34,243.74) and (165.22,245.86) .. (162.61,245.86) .. controls (159.99,245.86) and (157.87,243.74) .. (157.87,241.13) -- cycle ;
\draw  [color={rgb, 255:red, 255; green, 255; blue, 255 }  ,draw opacity=1 ][fill={rgb, 255:red, 208; green, 2; blue, 27 }  ,fill opacity=1 ] (211.37,310.12) .. controls (211.37,307.51) and (213.49,305.39) .. (216.1,305.39) .. controls (218.72,305.39) and (220.84,307.51) .. (220.84,310.12) .. controls (220.84,312.74) and (218.72,314.86) .. (216.1,314.86) .. controls (213.49,314.86) and (211.37,312.74) .. (211.37,310.12) -- cycle ;
\draw  [color={rgb, 255:red, 255; green, 255; blue, 255 }  ,draw opacity=1 ][fill={rgb, 255:red, 208; green, 2; blue, 27 }  ,fill opacity=1 ] (470.23,239.45) .. controls (470.23,236.83) and (472.35,234.71) .. (474.96,234.71) .. controls (477.58,234.71) and (479.7,236.83) .. (479.7,239.45) .. controls (479.7,242.06) and (477.58,244.18) .. (474.96,244.18) .. controls (472.35,244.18) and (470.23,242.06) .. (470.23,239.45) -- cycle ;
\draw [line width=0.75]    (422.99,73.53) -- (479.31,223.99) ;
\draw [shift={(480.01,225.86)}, rotate = 249.48] [color={rgb, 255:red, 0; green, 0; blue, 0 }  ][line width=0.75]    (10.93,-3.29) .. controls (6.95,-1.4) and (3.31,-0.3) .. (0,0) .. controls (3.31,0.3) and (6.95,1.4) .. (10.93,3.29)   ;
\draw [shift={(422.29,71.65)}, rotate = 69.48] [color={rgb, 255:red, 0; green, 0; blue, 0 }  ][line width=0.75]    (10.93,-3.29) .. controls (6.95,-1.4) and (3.31,-0.3) .. (0,0) .. controls (3.31,0.3) and (6.95,1.4) .. (10.93,3.29)   ;
\draw    (446.24,40.19) -- (414.56,76.22) ;
\draw    (428.9,56.7) .. controls (424.24,45.19) and (402.24,52.19) .. (395.24,66.19) ;
\draw  [color={rgb, 255:red, 255; green, 255; blue, 255 }  ,draw opacity=1 ][fill={rgb, 255:red, 208; green, 2; blue, 27 }  ,fill opacity=1 ] (409.83,76.22) .. controls (409.83,73.6) and (411.95,71.48) .. (414.56,71.48) .. controls (417.18,71.48) and (419.3,73.6) .. (419.3,76.22) .. controls (419.3,78.83) and (417.18,80.95) .. (414.56,80.95) .. controls (411.95,80.95) and (409.83,78.83) .. (409.83,76.22) -- cycle ;

\draw (362.46,288.06) node [anchor=north west][inner sep=0.75pt]  [rotate=-338.78]  {$a_{0}$};
\draw (499,142.4) node [anchor=north west][inner sep=0.75pt]    {$\alpha _{0}$};
\draw (225,246.4) node [anchor=north west][inner sep=0.75pt]    {$\textcolor[rgb]{0.74,0.06,0.88}{h}\textcolor[rgb]{0.74,0.06,0.88}{_{0}}$};
\draw (523,191.4) node [anchor=north west][inner sep=0.75pt]    {$\textcolor[rgb]{0.74,0.06,0.88}{h}\textcolor[rgb]{0.74,0.06,0.88}{_{1}}$};
\draw (379,38.4) node [anchor=north west][inner sep=0.75pt]  [color={rgb, 255:red, 189; green, 16; blue, 224 }  ,opacity=1 ]  {$\textcolor[rgb]{0.74,0.06,0.88}{h}\textcolor[rgb]{0.74,0.06,0.88}{_{2}}$};
\draw (143,168.4) node [anchor=north west][inner sep=0.75pt]    {$\textcolor[rgb]{0.74,0.06,0.88}{h}\textcolor[rgb]{0.74,0.06,0.88}{_{3}}$};
\draw (287,96.4) node [anchor=north west][inner sep=0.75pt]    {$\textcolor[rgb]{0.29,0.56,0.89}{s}$};
\draw (137,242.4) node [anchor=north west][inner sep=0.75pt]    {$\textcolor[rgb]{0.82,0.01,0.11}{D}$};
\draw (403,91.4) node [anchor=north west][inner sep=0.75pt]    {$\textcolor[rgb]{0.82,0.01,0.11}{C}$};
\draw (481.7,242.85) node [anchor=north west][inner sep=0.75pt]    {$\textcolor[rgb]{0.82,0.01,0.11}{B}$};
\draw (201,313.4) node [anchor=north west][inner sep=0.75pt]    {$\textcolor[rgb]{0.82,0.01,0.11}{A}$};
\draw (462.19,121.93) node [anchor=north west][inner sep=0.75pt]  [rotate=-47.94]  {$a_{1}$};
\draw (410,30.4) node [anchor=north west][inner sep=0.75pt]    {$\alpha _{1}$};

\end{tikzpicture}}
    \caption{Kinematics of the Bennett mechanism visualized as a skew isogram~\cite{bennett1914skew}. Rotation axes are $h_i$; DH parameters $a_i$ and $\alpha_i$; the line of symmetry is $s$.}
    \label{fig:bennet_kinematics}
\end{figure}

Overconstrained single-loop linkages are paradoxical movable linkages with n $\leq 6$  links, connected in series by 1-DoF joints. Such mechanisms are rigid in general, but may gain mobility because of special conditions on their design. Here we are interested in chains with revolute axes only. In case of $n=6$ the question of a full characterization of all such structures is still open, whereas all overconstrained linkages composed of 5 joints and links have been characterized by~\cite{Karger1998}. For $n=4$ there exists only one movable structure, the so-called Bennett mechanism~\cite{bennett1914skew}. Using Denavit-Hartenberg (DH)~\cite{Denavit1955} parameters the relations for mobility in the last case are given in Tab.~\ref{tab:dh_params_arbitrary}, where $a_i$ denotes the length of the common normal and $\alpha_i$ the twist angle between joints $i$ and $i+1$ and $d_i$ is the offset between the common normals of joint $i$ with the previous and proximate one. Geometrically $d_i=0, \, i=0 \dots 3$ means, that the common normals of the four joint axes have to form a skew isogram~\cite{bennett1914skew}, which is a spatial closed polygon, also shown in Fig.~\ref{fig:bennet_kinematics}. Finally, $\theta_i$ stands for the four joints variables. Furthermore, its opposite sides have equal lengths, opposite twist angles between adjacent joint axes have to be equal.

\begin{table}[pos=h]
    \caption{DH parameters of an arbitrary Bennett mechanism.}
    \label{tab:dh_params_arbitrary}
    \vspace{-10pt}
    \begin{center}
        \begin{tabular}{c | c c c c}
         i   &  $a_i$ &  $\alpha_i$ &  $d_i$ & $\theta_i$\\
         \hline\hline
         0 & $a_0$ & $\alpha_0$  & 0 & $\theta_0$\\
         \hline
         1 & $a_1$ & $\alpha_1$  &  0 & $\theta_1$\\
         \hline
         2 & $a_2=a_0$ & $\alpha_2=\alpha_0$  & 0 & $\theta_2$\\
         \hline
         3 & $a_3=a_1$ & $\alpha_3=\alpha_1$  & 0 & $\theta_3$\\
        \end{tabular}
    \end{center}
\end{table}

Additionally, these parameters have to fulfill the Bennett condition
\begin{equation}
    \frac{a_0}{\sin{\alpha_0}} = \frac{a_1}{\sin{\alpha_1}}\;.
\end{equation}

In this study, Bennett mechanisms with the design parameters in Tab.~\ref{tab:dh_params_ideal} were modeled and built. While alternative parameterizations exist for describing the kinematics of closed-chain mechanisms, we adopt DH parameters for their straightforward interpretability and suitability for direct comparison with the ideal structure. Since the identified errors are not intended for compensation in the control system, this choice does not impose additional constraints on the methodology.
Readers wishing to conduct their own analysis may do so using the experimental data provided, comparing, for example, the ideal and measured Plücker coordinates~\cite[p. 133]{Pottmann2001} or screw vectors~\cite{Zhao2018kinemat}. All data and algorithms used in this study are available as supplementary material~\cite{suppl_mat}.

\subsection{Recovery of Motion Polynomials from DH parameters}\label{sec:motion_recovery}

As we would like to have the tools of motion polynomials and the factorization theory of such polynomials at our disposal, we explain how to get from DH parameters, specifically the ones stated in Tab.~\ref{tab:dh_params_ideal} to the corresponding motion polynomial describing among other things the trajectory that the end-effector will travel along, and to which we can compare the simulations.

\begin{table}[pos=h]
    \caption{DH parameters of the 4R mechanism.}
    \label{tab:dh_params_ideal}
    \vspace{-10pt}
    \begin{center}
        \begin{tabular}{c | c c c c }
         i   &  $a_i$ [mm] &  $\alpha_i$ [deg] &  $d_i$ [mm] & $\theta_i$ [deg]\\
         \hline\hline
         0 & 110 & 150  & 0 & $\theta_0$\\
         \hline
         1 & 220 & 90  & 0 & $\theta_1$\\
         \hline
         2 & 110 & 150  & 0 & $\theta_2$\\
         \hline
         3 & 220 & 90  & 0 & $\theta_3$\\

        \end{tabular}
    \end{center}
\end{table}

We start by introducing the Denavit-Hartenberg transformation in their dual quaternion form, 
\begin{equation}
    \begin{gathered}
        r_x(\alpha_{i-1}) = \cos \frac{\alpha_{i-1}}{2} + \sin \frac{\alpha_{i-1}}{2} \mathbf{i} \; , \quad 
        t_x(a_{i-1}) = 1 - \varepsilon \frac{a_{i-1}}{2} \mathbf{i} \;, \\
        t_z(d_i) = 1 - \varepsilon \frac{d_i}{2} \mathbf{k} \;, \quad 
        r_z(\theta_i) = \cos \frac{\theta_i}{2} + \sin \frac{\theta_i}{2} \mathbf{k} \;,
    \end{gathered}
\end{equation}
where $\mathbf{i}$, $\mathbf{j}$, $\mathbf{k}$ are the quaternion basis units satisfying $\mathbf{i}^2 = \mathbf{j}^2 = \mathbf{k}^2 = \mathbf{ijk} = -1$, $\varepsilon$ is the dual unit with $\varepsilon^2 = 0$, and $\theta_i$ denotes the rotation angle of the $i$th revolute joint. They correspond to the angle of rotation about the common normal, the length of the common normal, the offset along the joint's rotational axis and the configuration dependent rotation angle about said axis respectively. They are known to completely describe the state of a chain of revolute axes. The transformation of the \((i-1)\)th axis to the \(i\)th axis is given\footnote{We are using the DH convention as in~\cite[Appendix C]{Lynch2017modernr}.} by
\begin{equation*}
    f_{i-1}^{i}(\theta_i) := r_x(\alpha_{i-1})t_x(a_{i-1})t_z(d_i)r_z(\theta_i) \;.
\end{equation*}
By fixing w.l.o.g. the first axis ($h_0$), we can get all other axes by using these transformations~\cite{huczala2022robkin}. To agree with the experimental setup above we set the axis \(h_0 = \mathbf{k}\) as the \(z\)-axis of the origin. Note that because of the closure of the linkage the axis $h_4$ is identical to $h_0$, and the axes $h_0$ and $h_3$ define the base (static) link.
For Tab.~\ref{tab:dh_params_ideal} we get\footnote{See supplementary material~\cite{suppl_mat} for details, file \textit{rational\textunderscore axes.py}.}
\begin{equation}
    \begin{gathered}
        h_0 = \mathbf{k} \; , \quad
        h_1 = \frac{\mathbf{j} - \sqrt{3}\mathbf{k}}{2} + \frac{11 \varepsilon}{200} (\sqrt{3} \mathbf{j} + \mathbf{k}) \; ,\\
        h_2 = \frac{\sqrt{3} \mathbf{j} + \mathbf{k}}{2} + \frac{11\varepsilon}{200}(\mathbf{j} - \sqrt{3} \mathbf{k}) \; , \quad
        h_3 = -\mathbf{j} +\frac{11}{50} \varepsilon \mathbf{k} \; .
    \end{gathered}
\end{equation}

These are Plücker coordinates of lines embedded in the dual quaternion space~\cite{huczala2024linkages}. A rotation around a line can be parameterized in dual quaternions by \(t - h\), where \(t \in \mathbb{P}^1\) denotes the cotangent of half the angle of joint angle $\theta$ and \(h\) is a vectorial unit dual quaternion, i.e.\ \(h + h^* = 0, hh^* = 1\).

Therefore, the motion of a serial chain of \(4\) rotational joints can be described by the product of these linear factors and the closure of a serial chain of rotational joints in terms of Denavit-Hartenberg parameters 
\begin{equation}
    [f_{3}^{0}(\theta_0) f_{0}^{1}(\theta_1) f_{1}^{2}(\theta_2) f_{2}^{3}(\theta_3)] = [1]
\end{equation}
can also be written in terms of the product of linear polynomials in the dual quaternions
\begin{equation} \label{eq:closure_condition}
    (t_0 - h_0) (t_1 - h_1) (t_2 - h_2) (t_3 - h_3) \in \mathbb{R} \setminus \{0\} \; .
\end{equation}
The set of parameter values \((t_0, t_1, t_2, t_3) \in (\mathbb{P}^1)^4\), where Eq.~\eqref{eq:closure_condition} holds is called \textit{configuration set} or \textit{configuration variety} of the mechanism. For our example in Tab.~\ref{tab:dh_params_ideal} we can solve this system of equations and obtain
\begin{equation}
    \left\{ (t_0, t_1, t_2, t_3) =\left(t, \sqrt{3} \, t, -t, -\sqrt{3} \, t\right) \mid t \in \mathbb{P}^1\right\}.
\end{equation}

Then, the motion of the third link (coupler) with respect to the base (static) link is given by the product \((t - h_0)(\sqrt{3} t - h_1)\). This is the motion polynomial and evaluates to
\begin{equation}
    \begin{aligned}
    \label{eq:motion}
    C(t) =
    &\sqrt{3}\, t^{2} +\frac{\sqrt{3}}{2}
    -\frac{\mathbf{i}}{2}
        -\frac{\mathbf{j}}{2} t-\frac{\sqrt{3}\, \mathbf{k}}{2} t \\
        &+\varepsilon\left(-\frac{11}{200}
        -\frac{11 \sqrt{3}\, \mathbf{i}}{200}
            -\frac{11 \sqrt{3}\, \mathbf{j}}{200} t-\frac{11 \mathbf{k}}{200} t
    \right).
    \end{aligned}
\end{equation}

\subsection{Flexible Multibody Modeling}\label{sec:flexible-mbs}

In the floating frame of reference formulation (FFRF)~\cite{Zwolfer2020, Zwoelfer2021},
the generalized coordinates of each flexible body are partitioned as
\begin{equation}
    \mathbf{q} = \begin{bmatrix} \mathbf{r}^T & \boldsymbol{\theta}^T & \mathbf{q}_f^T \end{bmatrix}^T,
    \label{eq:ffrf-coords}
\end{equation}
where $\mathbf{r} \in \mathbb{R}^3$ and $\boldsymbol{\theta} \in \mathbb{R}^3$
describe the position and orientation of a body-fixed reference frame,
and $\mathbf{q}_f$ contains the flexible deformation coordinates.
The equations of motion are assembled from the nodal mass and stiffness
matrices of the finite element model; see~\cite{Zwolfer2020} for the
full derivation.

\begin{figure*}[pos=tb]
    \centering
    \begin{subfigure}[t]{0.6\textwidth}
        \centering
        \includegraphics[width=\textwidth, trim={3mm 12mm 0mm 0mm}, clip]{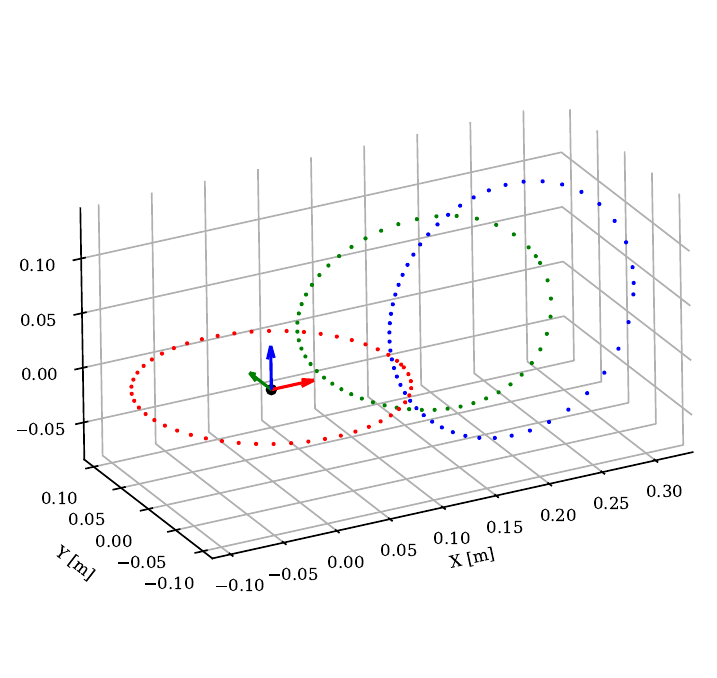}
        \caption{Orthogonal view.}
        \label{fig:vicon-path1}
    \end{subfigure}%
    \hfill
    \begin{subfigure}[t]{0.35\textwidth}
        \centering
        \includegraphics[width=\textwidth, trim={0mm -10mm 0mm 0mm}, clip]{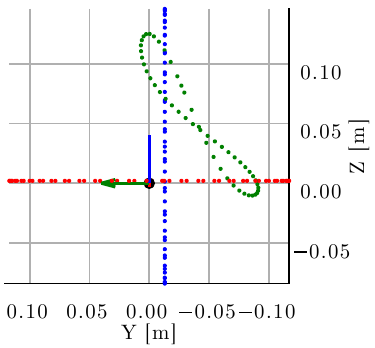}
        \caption{YZ plane projection view.}
        \label{fig:vicon-path2}
    \end{subfigure}
    \caption{Example of measured paths, traced by mechanism bodies; each path for each link. Black dot in origin -- base link (static); circular red and blue paths -- links attached to base link joints; green path -- coupler link (tool).}
    \label{fig:vicon-trajectories}
\end{figure*}

To keep the model computationally tractable, the flexible coordinates
are approximated by a truncated basis of Hurty/Craig-Bampton (HCB)
component modes~\cite{Hurty1965, Craig1968},
\begin{equation}
    \mathbf{q}_f \approx \boldsymbol{\Psi}\, \boldsymbol{\zeta},
    \label{eq:modal-reduction}
\end{equation}
where $\boldsymbol{\Psi}$ combines fixed-interface vibration modes and
constraint modes at the joint interfaces, and $\boldsymbol{\zeta}$ is the
reduced modal coordinate vector.
Both formulations are implemented in the open-source multibody code Exudyn~\cite{Exudyn2024}.
This code employs a constrained multibody formulation, which can handle serial and parallel mechanisms equally well. The equations of motion in Exudyn are represented by the ordinary differential equations of the rigid and flexible bodies together with algebraic joint constraints. The resulting index-3 differential-algebraic equations are solved with the generalized-$\alpha$ solver.
The concrete simulation parameters are given in Sect.~\ref{sec:accuracy-analysis}.

Flexible modeling is essential for this study for two reasons.
First, the tested materials (PLA, cardboard, and bamboo) exhibit finite
bending stiffness that cannot be represented by a rigid-body model.
Second, the Bennett mechanism is an overconstrained linkage whose
rigid-body formulation contains redundant constraints. As discussed for
multibody systems with redundant constraints~\cite{GarcadeJaln2013, Frczek2011}, such
systems may lead to singular or rank-deficient constraint Jacobians
unless the redundancy is treated explicitly. In a purely rigid geometric
model, deviations from the exact Bennett conditions generally eliminate
the theoretical mobility and may render assembly impossible. In contrast,
structural compliance relaxes the redundant constraints through elastic
deformation, allowing the mechanism to remain mobile despite manufacturing
imperfections, consistent with the behavior observed in the physical
prototypes.

\subsection{Kinematic Parameters Identification}\label{sec:identification}

This subsection describes the identification process for obtaining the DH parameters of single-loop structures, which are later used to compare the assembly accuracy of the physical mechanisms against their simulated counterparts. The methodology requires tracking all links of the mechanism throughout its motion, which was achieved using a camera-based motion capture system, specifically, the Vicon system described in Sect.~\ref{sec:setup-measurements}.

The prerequisite is the acquisition of body trajectories, as shown in Fig.~\ref{fig:vicon-trajectories}. At least one point shall be measured on each body; however, as explained in more detail in Sect.~\ref{sec:setup-measurements}, we attached multiple markers to each body, from which the Vicon system computed the 6-DoF body pose, increasing trajectory accuracy, as described in~\cite{Merriaux2017}. In every instance $t_i$ from $i \in 0 \dots n$ ($n$ is the number of measured positions of each marker), the position points $P_{ji} \in \mathbb{R}^3$ of each body $j$ were recorded. The rotational axes of the joints have to be identified first. 

The algorithm is as follows: divide a single-loop linkage in two branches (they start in the base frame that is grounded and meet again in the tool link). For each branch, identify the axes separately, starting from the base link in the direction of the coupler link. The axes are obtained as lines in space in a single specific configuration of the mechanisms. 

Finding the first joint axis is trivial as the path is already circular; therefore, 
PCA via SVD
\cite{Mirkin2011} is applied to the mean-centered path coordinates. 
Let $\mathbf{X} \in \mathbb{R}^{n \times 3}$ denote the $n$ recorded 
end-effector positions, and let
\begin{equation}
    x_c = \frac{1}{n}\sum_{i=1}^{n} x_i
\end{equation}
be the centroid. The centered matrix is then decomposed as
\begin{equation}
    \tilde{\mathbf{X}} = \mathbf{U}\mathbf{S}\mathbf{V}^T \; ,
\end{equation}
where $\mathbf{U} \in \mathbb{R}^{n \times n}$, 
$\mathbf{S} \in \mathbb{R}^{n \times 3}$, and 
$\mathbf{V} \in \mathbb{R}^{3 \times 3}$.
PCA transforms the data into a new coordinate frame based on variance by its scalar projection. Thus, the columns of matrix $\mathbf{V}$ correspond to the coordinate frame~\cite{Mirkin2011}. The first two columns are basis vectors in the plane that approximates the circular path, and the last column is the vector with least variance, i.e., the searched direction of joint axis $\vec{a} \in \mathbb{R}^3$.

\begin{figure}[pos=b]
    \centering
    \includegraphics[width=\linewidth]{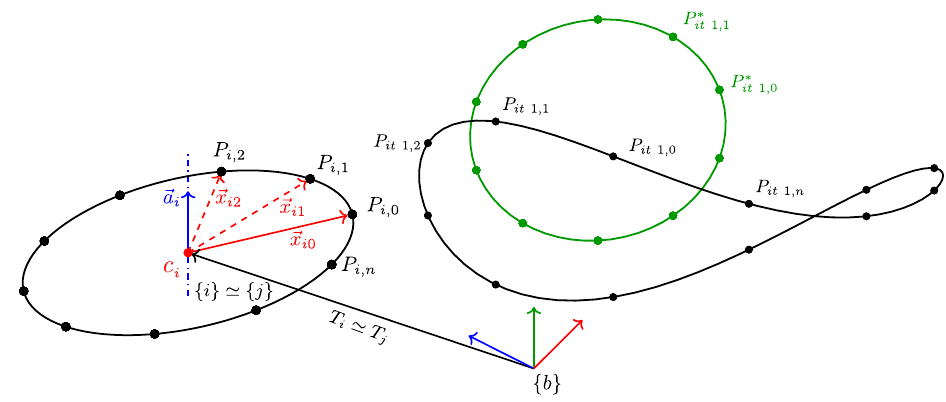}
    \caption{Projection of the coupler quadratic curve path in the previous link with relative rotation.}
    \label{fig:projection}
\end{figure}

To define a joint axis fully, a point on the axis is needed. Therefore, in a subsequent step, we acquire the point $c \in \mathbb{R}^3$, which is the approximate circle-center of the measured circular path points. The path points are projected in a best fitted 2D plane, where a linear least-squares problem is solved in this 2D space to obtain the circle center $(c_x, c_y) \in \mathbb{R}^2$, and finally the 3D circle center is reconstructed as 
\begin{equation}
    c = x_c + c_x \vec{e}_x + c_y \vec{e}_y
\end{equation}
where $\vec{e}_x$ and $\vec{e}_y \in \mathbb{R}^3$ are orthonormal bases of the 2D plane. The output, axis direction $\vec{a}$ and a point on that axis $c$ are used to calculate the Pl{\"u}cker coordinates of a line~\cite[p. 133]{Pottmann2001}.

As was shown in Fig.~\ref{fig:vicon-trajectories}, the trajectory of the coupler is no longer circular. However, a projection into the moving frame of a previous joint is feasible, if the measured points are ordered (or if the measuring system captures the points with a timestep). The projection is visualized in Fig.~\ref{fig:projection}.

The origin of the measurements is the base frame $\{b\}$ with transformation $\mathbf{T}_b$ equal to identity. We construct the transformation $\mathbf{T}_i$, which will remain constant, from previously determined vectors $\vec{a}_i, \vec{x}_{i0}$ and center $c_i$ with $i$ marking the index of the previously determined joint.
A homogeneous point $p_{i+1,j}$, where $j = 0,1 \dots n$ is the number of measured points, is then projected as
\begin{equation}
    p_{i+1,j}^* = \mathbf{T}_i \mathbf{T}_{i,j}^{-1} p_{i+1,j}
    \label{eq:projection}
\end{equation}
with $\mathbf{T}_{i,j}$ being the transformation matrix that rotates with the link $i$ in respect to origin $\mathbf{T}_b$, and is constructed from $c_i, \vec{a}_i$ and variable $\vec{x}_{i,j}$. The set of projected points $p_{i+1,j}^*$ then appears as circular-like path in $\mathbf{T}_b$, and the PCA related methodology above can be performed. 


In this way, the lines representing the mechanism axes are obtained, i.e. the kinematic parameters identification process is finalized. Note that Eq.~\eqref{eq:projection} can be extended from the left, for example by pair $\mathbf{T}_{i-1} \mathbf{T}_{i-1,j}^{-1}$, if the linkage branch has three revolute joints.
The DH parameters are obtained from the Pl{\"u}cker coordinates using the algorithm in~\cite{huczala2022robkin} implemented in the Rational Linkages package~\cite{huczala2024linkages}.

\begin{figure*}[pos=tbp]
    \centering
    \begin{subfigure}[t]{0.4\textwidth}
        \centering
        
        \includegraphics[width=\textwidth, trim={0mm 0mm 0mm 0mm}, clip]{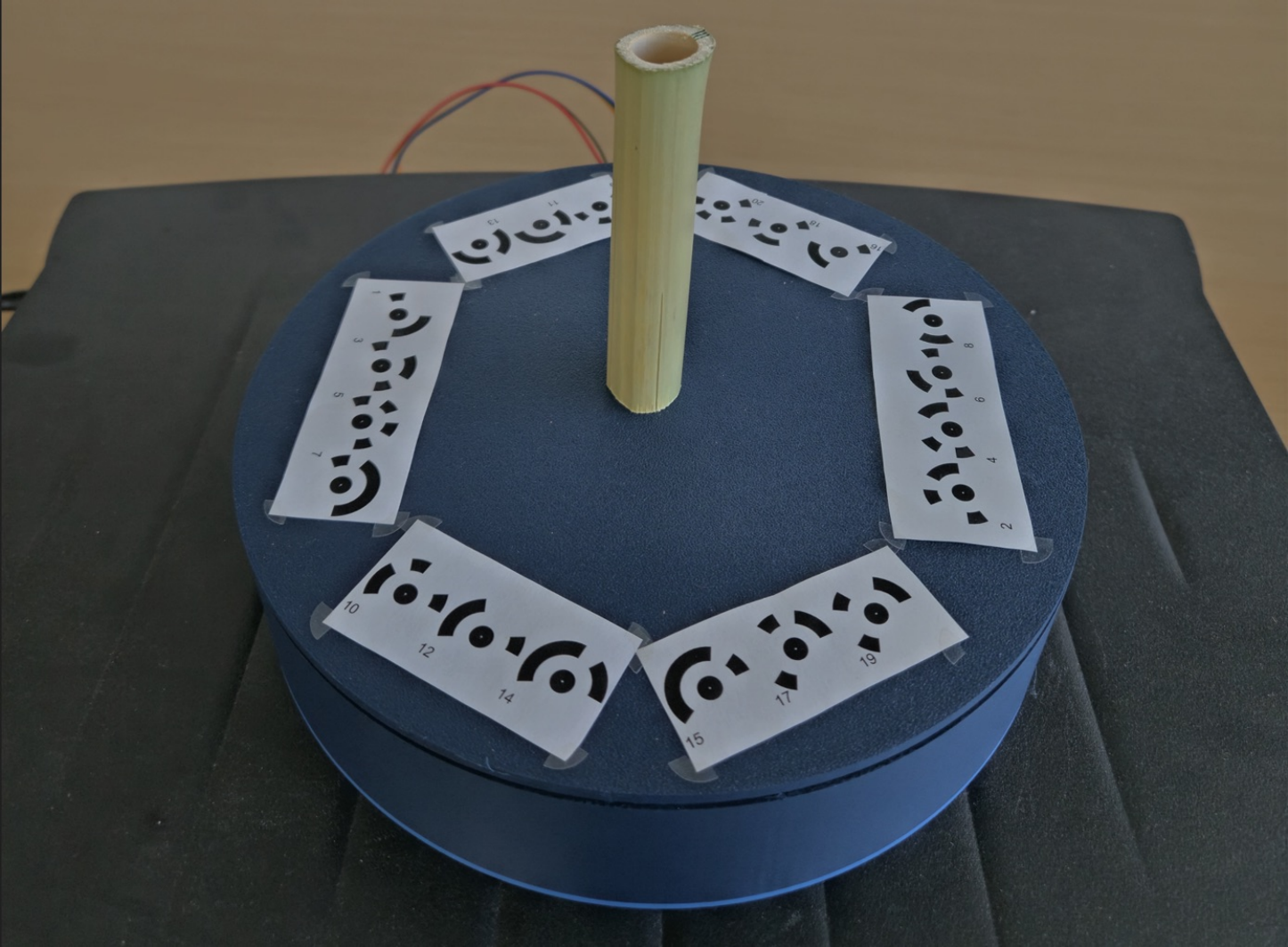}
        
        \caption{}
        \label{fig:bamb_scanning_setup}
    \end{subfigure}%
    \hfill
    \begin{subfigure}[t]{0.4\textwidth}
        \centering
        
        \includegraphics[width=\textwidth, trim={0mm 0mm 0mm 0mm}, clip]{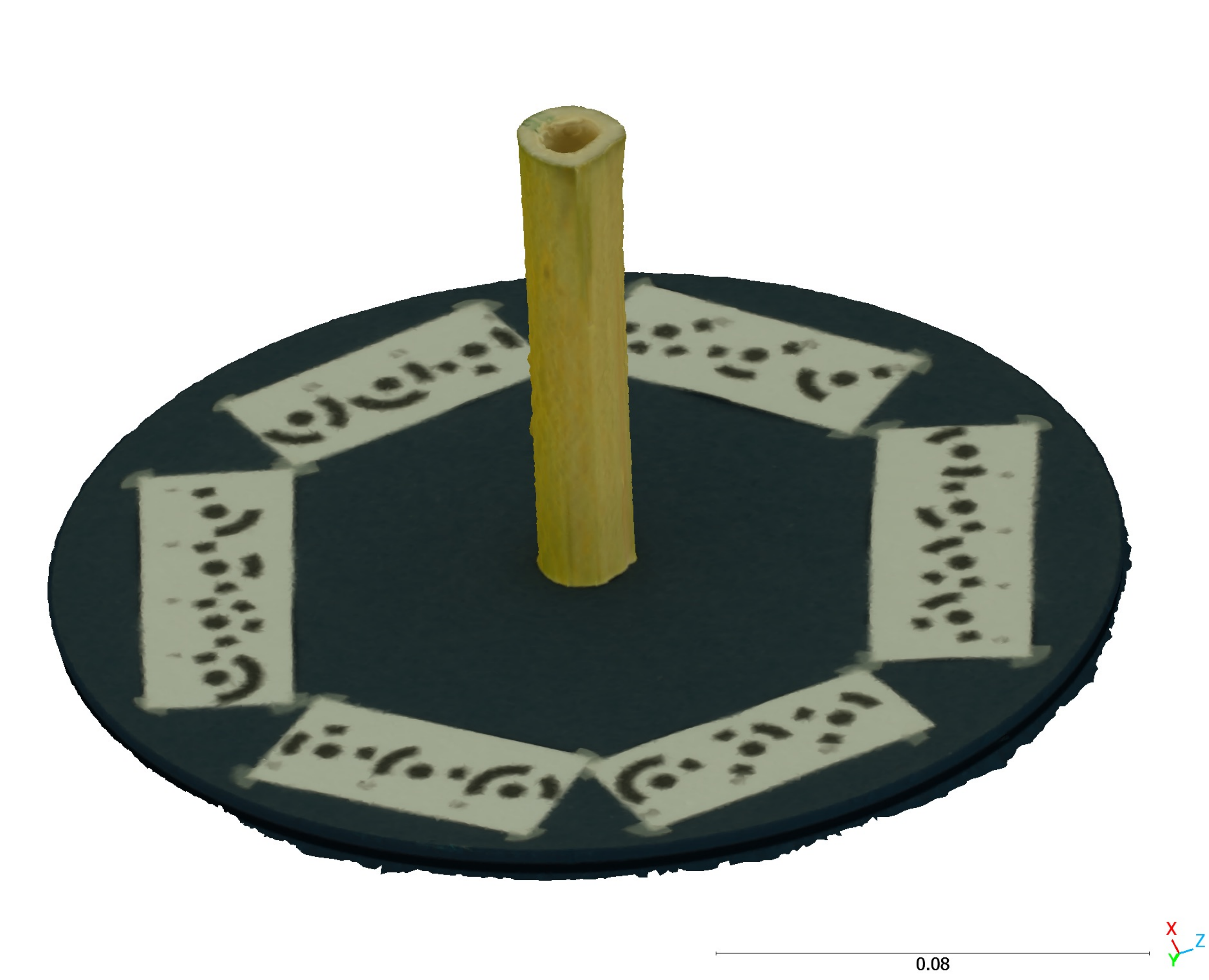}
        
        \caption{}
        \label{fig:bamb_meshing}
    \end{subfigure}
    \caption{(a) Photogrammetric scanning setup. (b) Mesh derived from photogrammetry.}
    \label{fig:bambo_scan_mesh}
\end{figure*}

\subsection{Bamboo Scanning and 3D Model Meshing}\label{sec:bamboo-scanning}
To facilitate the mutual adaptation of the bamboo components, which will be used as links in Sect.~\ref{sec:assembly_of_mechs}, digital reconstruction via 3D meshing was required. Such 3D model of the bamboo scan was integrated into the design, as shown later in Fig.~\ref{fig:bamb_model}.

The objective of this step was to generate a geometrically accurate mesh obtained at low cost. A photogrammetric approach was selected, utilizing a smartphone (Samsung Galaxy S23) and the free-to-use software \textit{RealityScan}.
To ensure robust acquisition, the smartphone was mounted on a tripod while each bamboo was placed on a 3D-printed turntable, see Fig.~\ref{fig:bambo_scan_mesh}, separately. Images were captured at regular intervals from two elevation angles. To facilitate accurate scaling during post-processing, 12-bit coded targets were affixed to the turntable, and the relative distances were measured. Given the weakly textured surface of the bamboo, the smartphone’s ``Pro mode’’ was employed, to ensure an optimal image quality. In this mode parameters such as manual focus, ISO, aperture, and exposure time were optimized and fixed to ensure consistency across the captured images, so that in the end, about 70 images were taken of each object on the rotating table.

A Structure from Motion (SfM) workflow~\cite{Schoenberger2016} was applied to calculate the intrinsic and extrinsic camera parameters. To prevent feature conflict between the static background and the rotating object, image-masks were applied to isolate the bamboo components in every image. A bundle adjustment~\cite{triggs2000} was subsequently performed to optimize parameters and enforce the scale constraints defined by the coded targets. A dense point cloud was then computed using a Multi-View Stereo algorithm~\cite{furukawa2015} and triangulated into a 3D mesh. The resulting meshes averaged $3 \times 10^5$ faces, corresponding to a resolution of approximately $\rm 0.1~mm$.


\begin{figure}[pos=b]
    \centering
    \includegraphics[width=\linewidth]{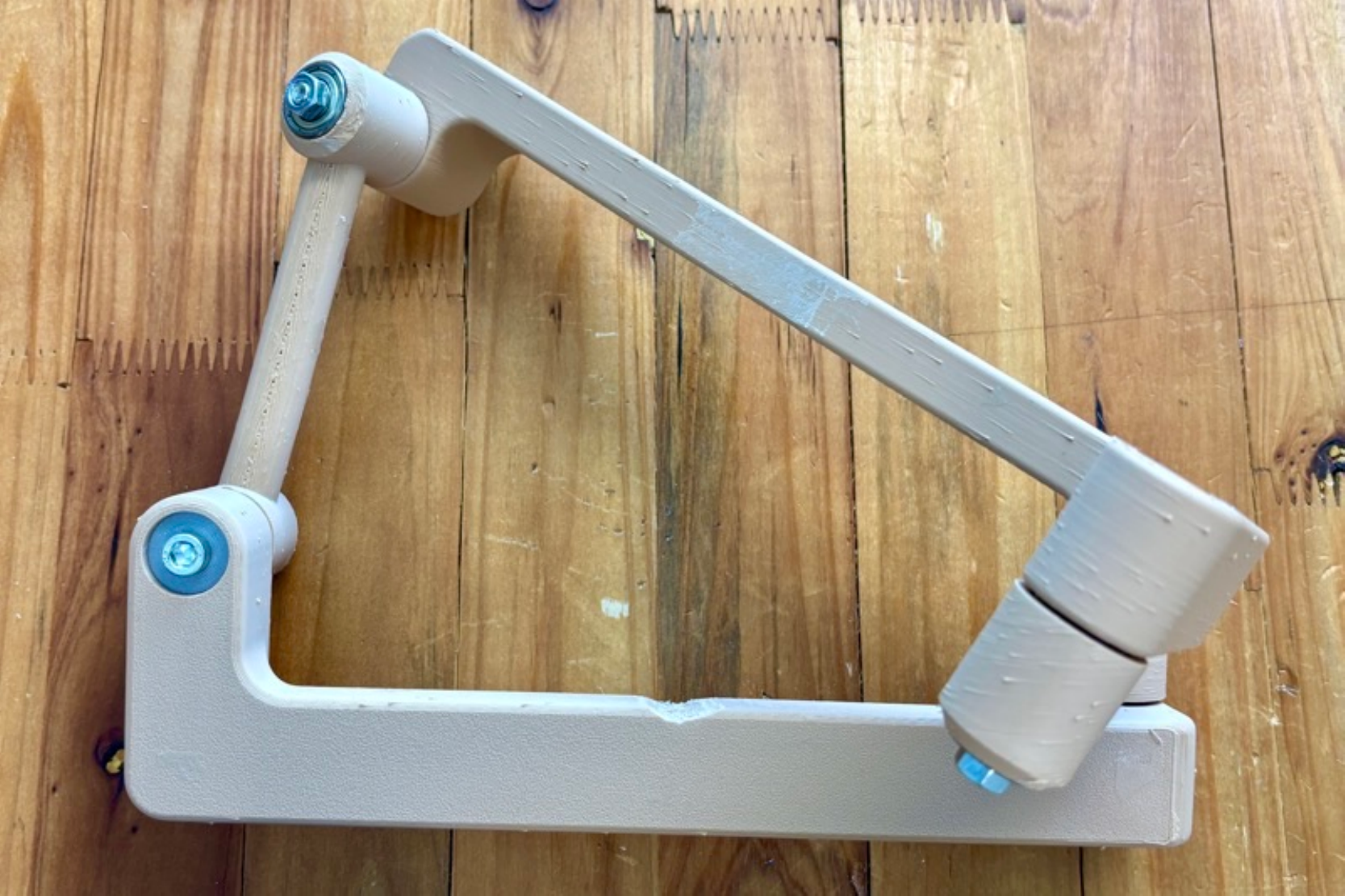}
    \caption{3D printed PLA prototype.}
    \label{fig:plastic_assembly}
\end{figure}


\section{Mechanisms Design and Assembly}\label{sec:assembly_of_mechs}

This section describes the mechanisms built during this study. Initially, the scope was limited to reporting findings on the fully 3D-printed prototype to support the flexible modeling approach presented for overconstrained linkages. However, experiments with alternative materials revealed interesting and unexpected properties. Although further research is needed in this direction, we believe that reporting on these findings, with cardboard tubes and bamboo sticks as building blocks, offers value to the community and may encourage exploration of unconventional, sustainable design practices.

The first prototype, shown in Fig.~\ref{fig:plastic_assembly}, consists of 3D-printed links from the standard PLA material. The joints consist of pairs of standard bearings of type 626, connected with precise shoulder screws. 

The second prototype, shown in Fig.~\ref{fig:paper_assembly}, is made of cardboard tubes, which are glued to the 3D-printed joints. The tubes have 25 mm outer diameter and 22 mm inner diameter.
This lightweight design benefits from the straightforward assembly (the tubes need only be cut to a specific length) and symmetric geometry of the tubes. At this scale, they provide high durability. From a basic usage and motion performance point of view, no difference was observed compared to the plastic prototype.

\begin{figure*}[pos=tb]
    \centering
    \begin{subfigure}{0.3\textwidth}
        \centering
        \includegraphics[width=\textwidth]{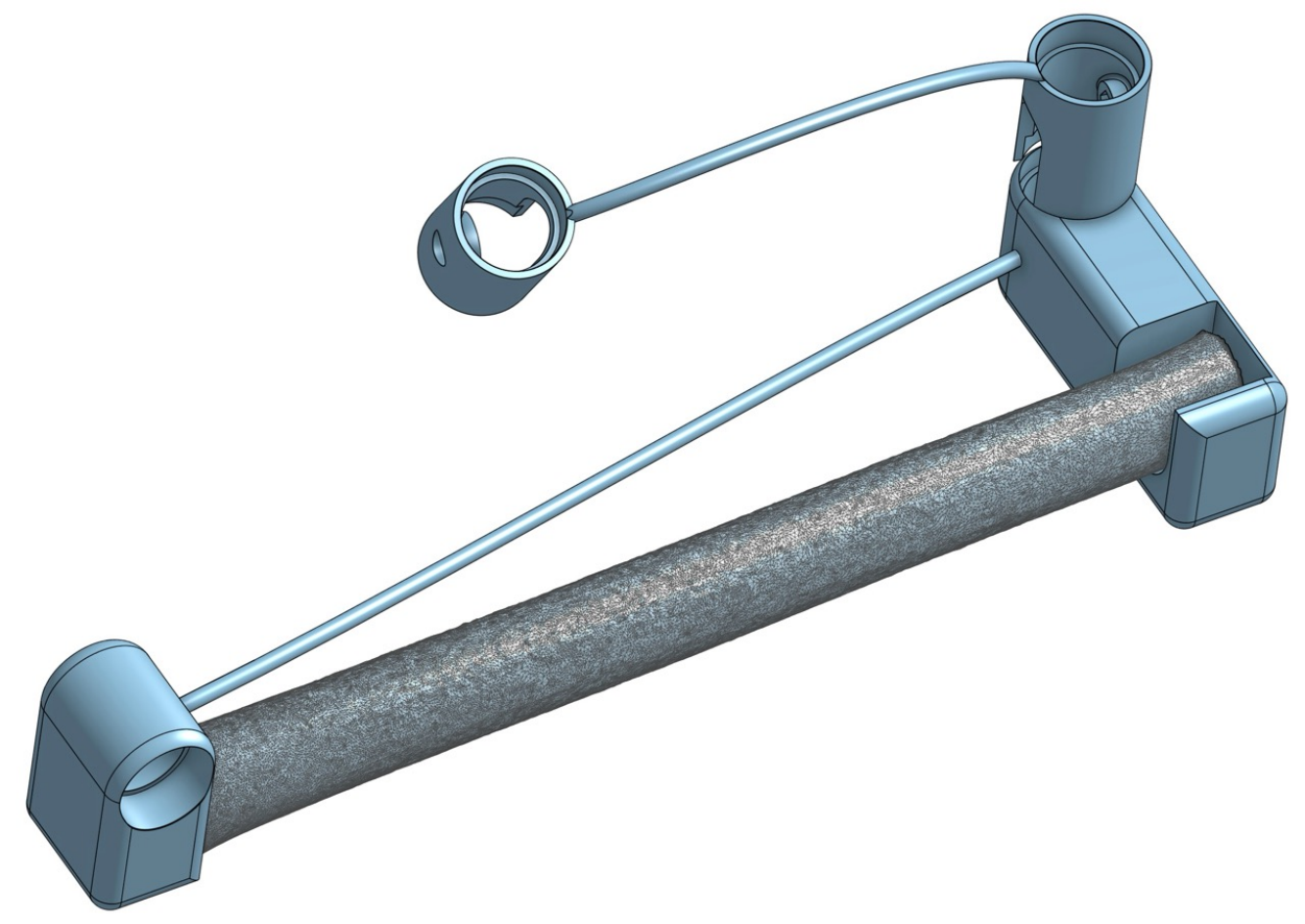}
        \caption{3D model of bamboo links.}
        \label{fig:bamb_model}
    \end{subfigure}%
    \hfill
    \begin{subfigure}{0.355\textwidth}
        \centering
        \includegraphics[width=\textwidth]{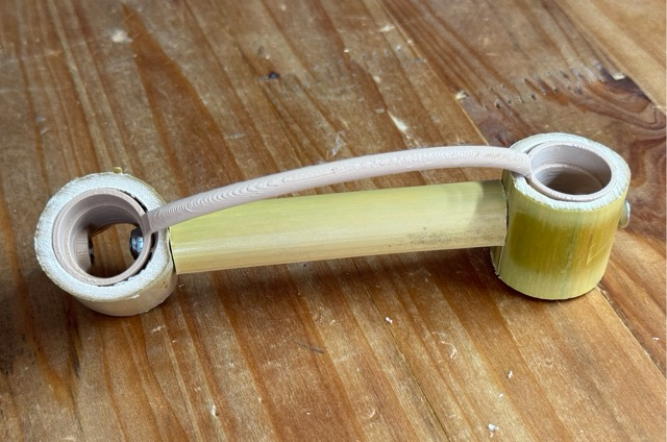}
        \caption{Insert fitting in the bamboo link.}
        \label{fig:bamb_insert_in}
    \end{subfigure}%
    \hfill
    \begin{subfigure}{0.33\textwidth}
        \centering
        \includegraphics[width=\textwidth]{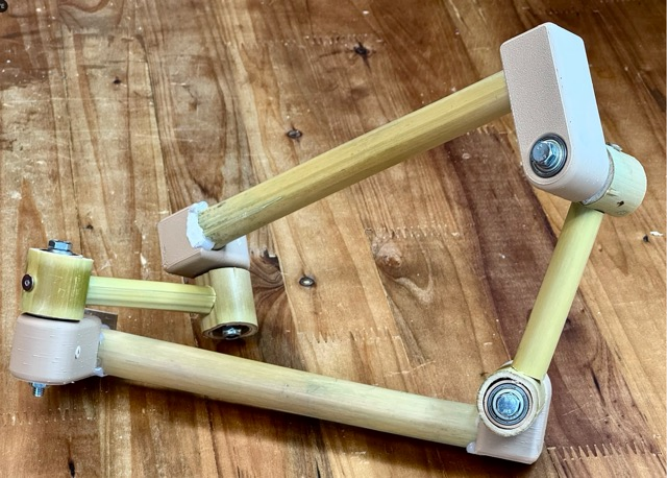}
        \caption{Bamboo linkage assembly.}
        \label{fig:bamb_assembly}
    \end{subfigure}
    \caption{Design and assembly of the bamboo linkages. The 3D printed insert was designed to fit into the bamboo stick and provide an accurate fitting according to DH parameters of the joints.}
    \label{fig:bamboo_mech}
\end{figure*}

\begin{figure}[pos=b]
    \centering
    \includegraphics[width=\linewidth]{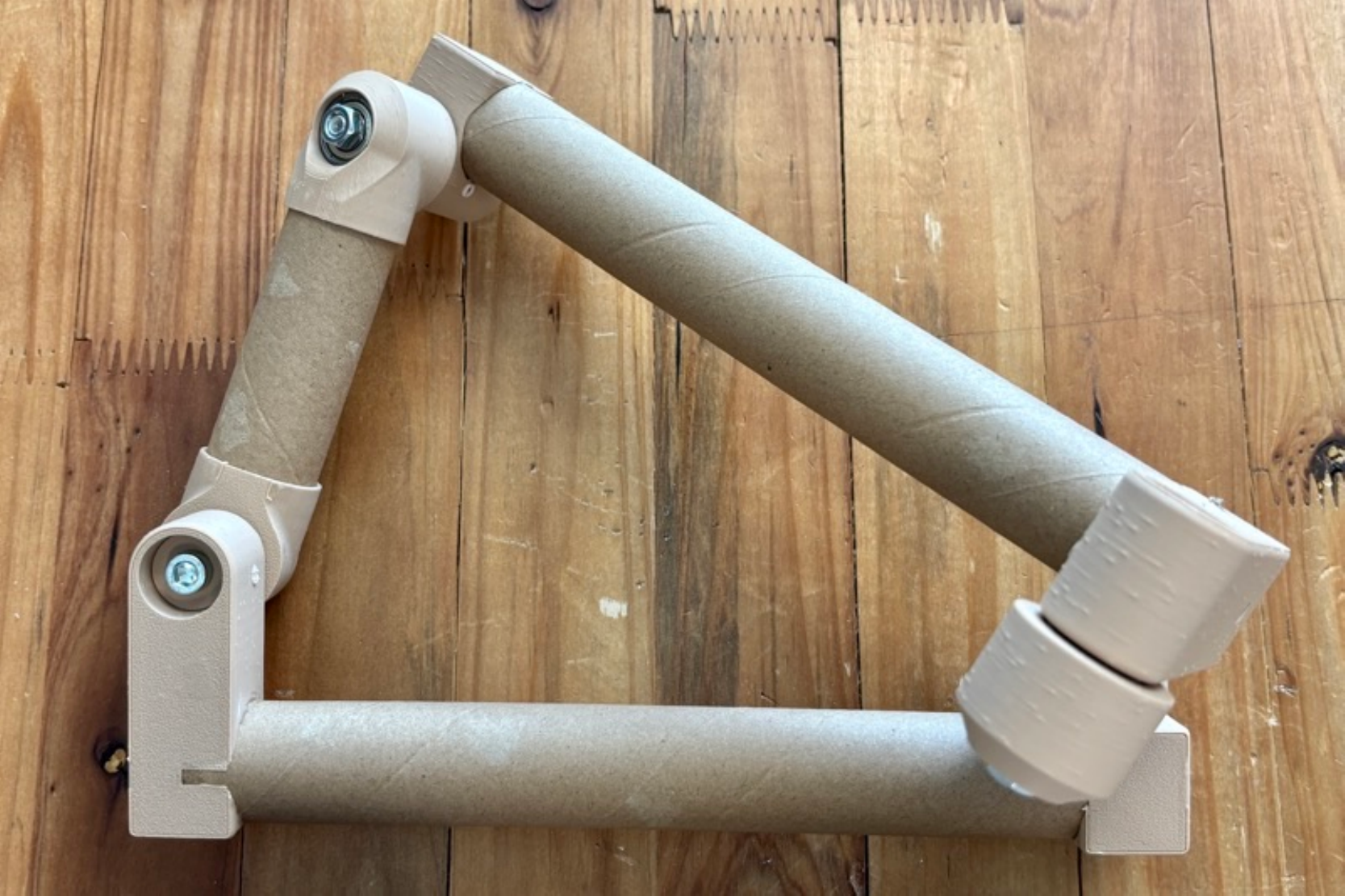}
    \caption{Assembled cardboard linkage.}
    \label{fig:paper_assembly}
\end{figure}

The third prototype, shown in Fig.~\ref{fig:bamboo_mech}, is made of bamboo sticks. The assembly was a challenge due to the natural variability of the bamboo rods. To overcome this issue, a 3D-printed insert was designed to fit the bamboo stick and provide an accurate position according to the DH parameters of the joints. 
The modeling was performed using the scanning methodology described in Sect.~\ref{sec:bamboo-scanning}, which was especially crucial to establish the connection between the bamboo stick and the 3D-printed joint sockets. 

The cardboard and bamboo prototypes introduced natural assembly misalignments; however, they still performed the desired motion without any issues, although the Bennett condition was not (from the strict mathematical point of view) satisfied. Therefore, we decided to introduce intentional misalignments in two other 3D-printed plastic prototypes to compare the results and validate the proposed flexible modeling approach. The base link was therefore altered by $1^\circ$ and $2^\circ$ changing the parameter $\alpha_3 = 91^\circ$ and $92^\circ$ in two different variants of the plastic prototype. The results are later compared to simulation in Sect.~\ref{sec:results}.

\section{Accuracy Analysis}\label{sec:accuracy-analysis}

This section covers simulation and laboratory setup to compare the modeling approach with experiments. 

\subsection{Simulation Setup}

The flexible multibody simulation was carried out in Exudyn (used version: 1.11.0), as shown in Fig.~\ref{fig:exu_setup}.
Each of the four links is modeled as an independent elastic body using the FFRF (see Sect.~\ref{sec:flexible-mbs}), in which the total motion is decomposed into a large rigid-body displacement described by the floating frame and a superimposed small elastic deformation. To keep the system computationally tractable, the elastic degrees of freedom of each link are reduced via the HCB substructuring method, shown in Eq.~\eqref{eq:modal-reduction}. In this approach, the interface regions at both ends of a link, corresponding to the two bearing bores, serve as boundary sets, and the interior dynamics are represented by a truncated set of $n_m = 32$ fixed interface normal modes. The underlying finite element meshes were generated from the STL geometry of the physical links using linear tetrahedral elements with a maximum edge length of $2$~mm.

Links are assumed to be isotropic and linearly elastic. The PLA prototypes were printed with three perimeters and 15\% infill, which reduces the effective stiffness and density compared to bulk PLA~\cite{Lubombo2018effecto, Farazin2021effecto}; the simulation used $E = 1.2$~GPa, $\nu = 0.33$, and $\rho = 450$~kg/m$^3$.
An ideally rigid reference case was obtained by setting $E = 10^{15}$~Pa and $\rho = 1250$~kg/m$^3$, rendering elastic deformations negligible while keeping the same model structure. Small stiffness-proportional (Rayleigh) damping with coefficient $\beta = 0.005$ is included to dissipate high-frequency elastic oscillations.

Each joint, see Fig.~\ref{fig:exu_setup}, is an ideal revolute joint, and the coupling between the revolute joint and the deformable link is established by rigidly averaging the kinematics of the finite element nodes located at the cylindrical bearing surface, so that the joint acts on a weighted mean position and orientation of that interface region. Link 0 (base) is rigidly clamped at its mounting faces, suppressing all translational and rotational degrees of freedom at that location.

The mechanism is set in motion by prescribing the angular velocity at the driven joint adjacent to the base. To avoid impulsive loading, the angular velocity is increased smoothly from rest to a steady-state value of $\omega = 1$~rpm following a cosine profile over a ramp interval of $t_\mathrm{ramp} = 1$~s, after which the mechanism completes exactly one full revolution. Gravitational acceleration $\mathbf{g} = [0,\, 0,\, {-9.81}]$~m/s$^2$ is applied throughout.

To assess the influence of manufacturing tolerances, several simulation variants were evaluated: (i)~an ideal rigid model with nominal DH parameters, (ii)~the flexible PLA model with nominal link lengths, and (iii)~flexible models with an intentional joint-axis misalignment of $1^\circ$ and $2^\circ$ introduced at joint 3. For each variant, three surface nodes per link were tracked on link 1, link 2, and link 3 to define a local body frame, mirroring the Vicon rigid body cluster approach; link 0 (base) is fixed and thus excluded. Since simulation data is noise-free, three points suffice to uniquely determine the frame. The resulting trajectories were subsequently processed with the kinematic identification procedure described in Sect.~\ref{sec:identification}, providing a direct comparison to the physical measurements.

\begin{figure*}[pos=tbp]
    \centering
    \begin{subfigure}[t]{0.33\textwidth}
        \centering
        \includegraphics[width=\textwidth, trim={30mm 0mm 40mm 0mm}, clip]{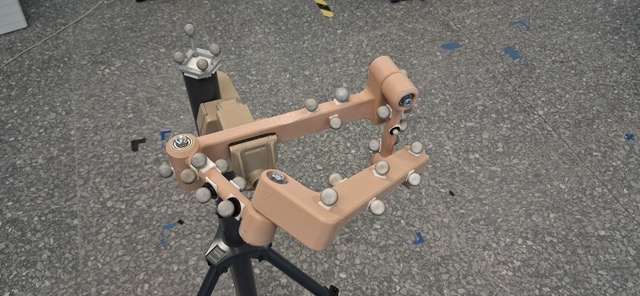}
        \caption{PLA prototype}
        \label{fig:plastic}
    \end{subfigure}%
    \hfill
    \begin{subfigure}[t]{0.33\textwidth}
        \centering
        \includegraphics[width=\textwidth, trim={30mm 0mm 40mm 0mm}, clip]{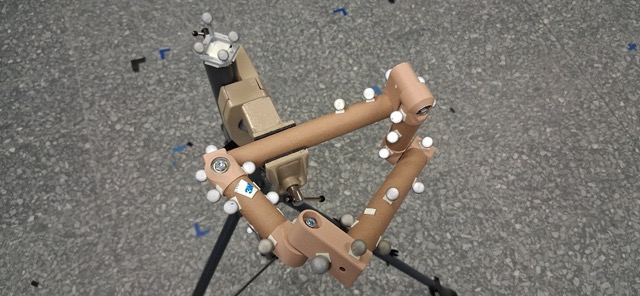}
        \caption{Cardboard prototype}
        \label{fig:paper}
    \end{subfigure}%
    \hfill
    \begin{subfigure}[t]{0.33\textwidth}
        \centering
        \includegraphics[width=\textwidth, trim={35mm 0mm 35mm 0mm}, clip]{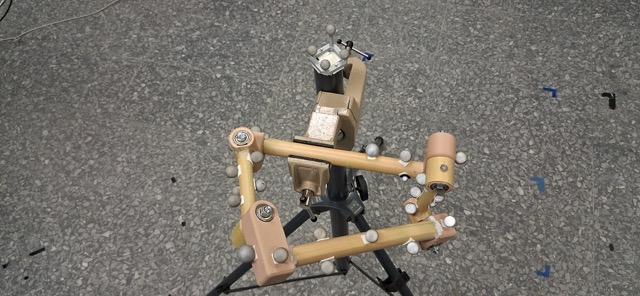}
        \caption{Bamboo prototype}
        \label{fig:bamboo}
    \end{subfigure}
    \caption{Laboratory setup of the three prototypes.}
    \label{fig:measured_mechanisms}
\end{figure*}

\begin{figure}[pos=b]
    \centering
    \includegraphics[width=\linewidth]{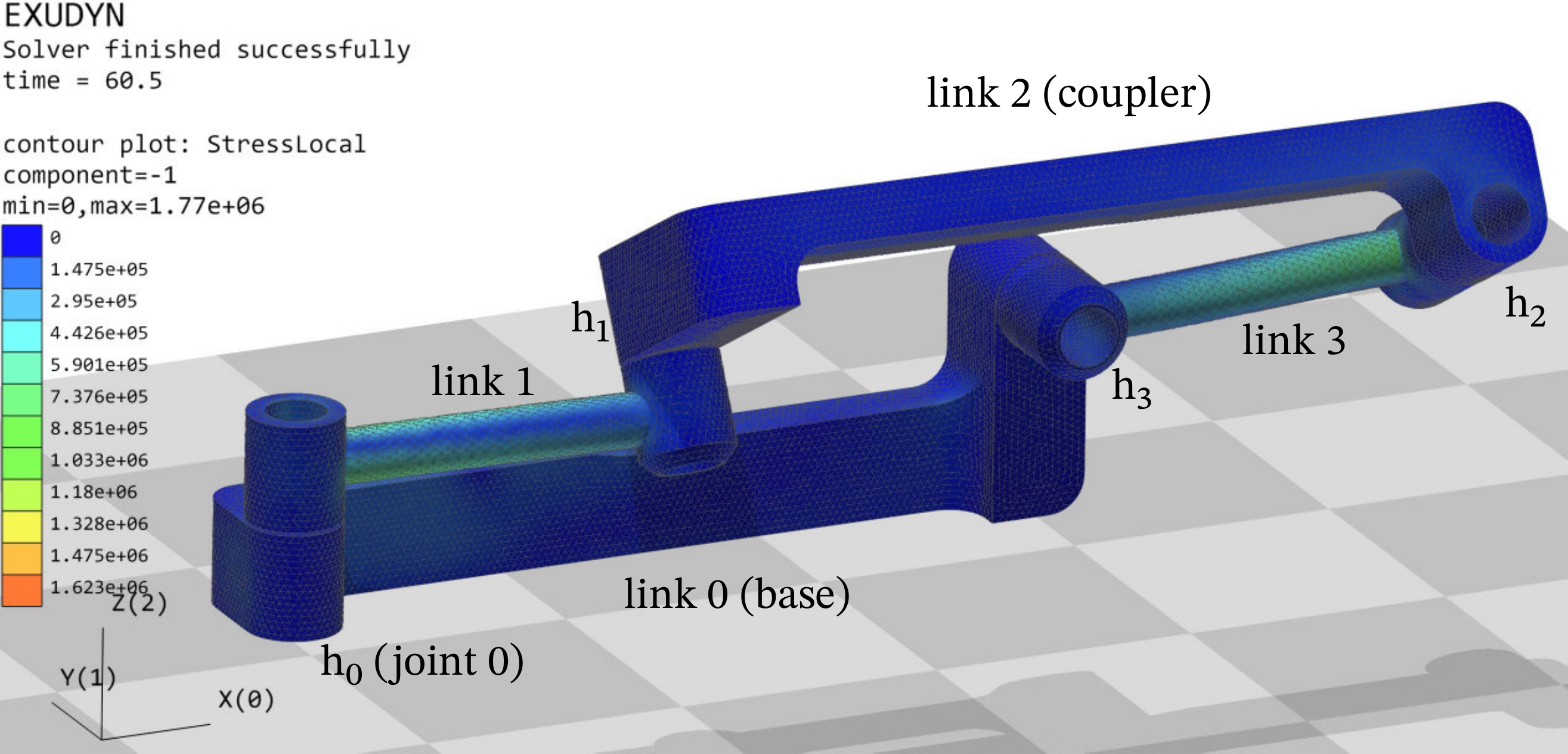}
    \caption{Meshed flexible multibody model of the four-bar linkage in Exudyn.}
    \label{fig:exu_setup}
\end{figure}

\subsection{Setup and Measurements}\label{sec:setup-measurements}

The physical experiments were carried out using a Vicon optical motion-capture system~\cite{Merriaux2017}. Rigid marker clusters were attached to each of the four links (link 0 -- link 3) and to a fixed reference frame (origin) mounted near the base joint, as shown in Fig.~\ref{fig:measured_mechanisms}. For each captured frame the system outputs the full 6-DoF pose (translations $T_X, T_Y, T_Z$ and Euler angles $R_X, R_Y, R_Z$) of every marker cluster. Each prototype was manually moved through one full revolution. Since the comparison is based on geometric trajectory analysis and DH parameter identification rather than time-domain matching, the variable speed of manual actuation does not affect the validity of the comparison. Five independent runs were recorded for each prototype configuration: the nominal PLA prototype, the PLA prototype with an intentionally introduced joint-axis misalignment of $1^\circ$ and $2^\circ$, the cardboard prototype, and the bamboo prototype, yielding a total of 25 measurement sequences.

Raw capture data were preprocessed by applying a centered moving-average filter with a window of 10~frames to suppress high-frequency marker noise. The 3-D translation trajectories of three links (link 1, link 2, link 3) were then used as input to the kinematic identification procedure described in Sect.~\ref{sec:identification}: a 3-D circle is fitted to each trajectory by first projecting the points onto the best-fit plane via SVD and then solving a linear least-squares problem in the plane. The circle center and its normal vector define the joint axis direction, from which the DH parameters $a_i$, $\alpha_i$, and $d_i$ are extracted over all recorded revolutions. The mean and standard deviation reported in Tab.~\ref{tab:dh_deviation} for laboratory experiments are computed across the five runs of each prototype.

\section{Results}\label{sec:results}

\begin{figure*}[pos=tbp]
    \centering
    \begin{subfigure}[t]{0.55\textwidth}
        \centering
        \includegraphics[width=\textwidth, trim={40mm 32mm 10mm 35mm}, clip]{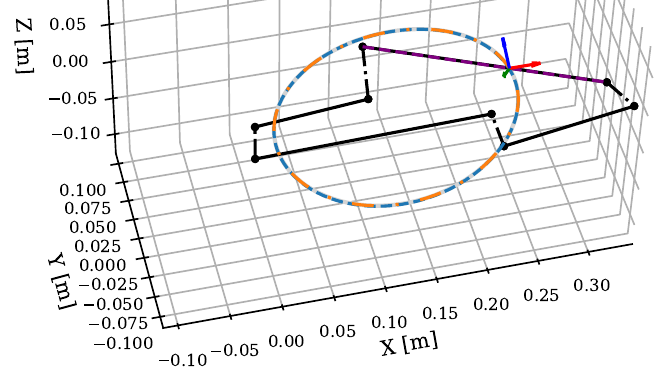}
        \caption{Overall view on the ideal and simulated trajectories.}
        \label{fig:comparison}
    \end{subfigure}%
    \hfill
    \begin{subfigure}[t]{0.44\textwidth}
        \centering
        \includegraphics[width=\textwidth, trim={0mm -18mm 0mm 0mm}, clip]{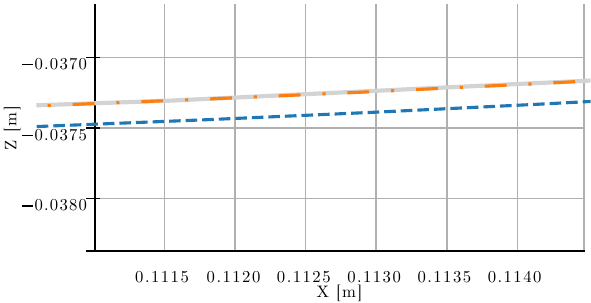}
        \caption{Detailed comparison in XZ plane projection.}
        \label{fig:comparison_detail}
    \end{subfigure}
    \caption{Comparison of the ideal and simulated trajectories. Black: line model of the linkage; gray: ideal trajectory (motion curve); orange: simulated trajectory with stiff material; blue: simulated trajectory with flexible material set to PLA properties.}
    \label{fig:compare_trajs}
\end{figure*}

\begin{figure*}[pos=tbp]
    \centering
    \begin{subfigure}[t]{0.49\textwidth}
        \centering
        \includegraphics[width=\textwidth]{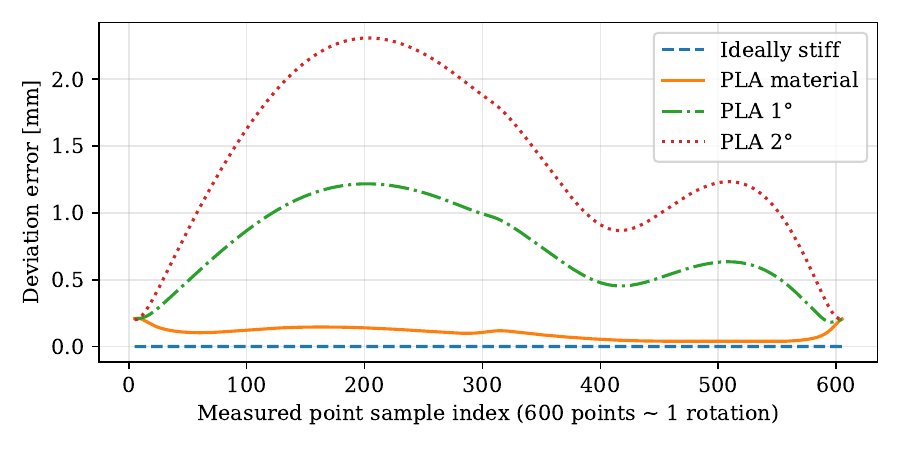}
        \caption{Deviation error between simulated trajectories.}
        \label{fig:deviation_err}
    \end{subfigure}%
    \hfill
    \begin{subfigure}[t]{0.49\textwidth}
        \centering
        \includegraphics[width=\textwidth]{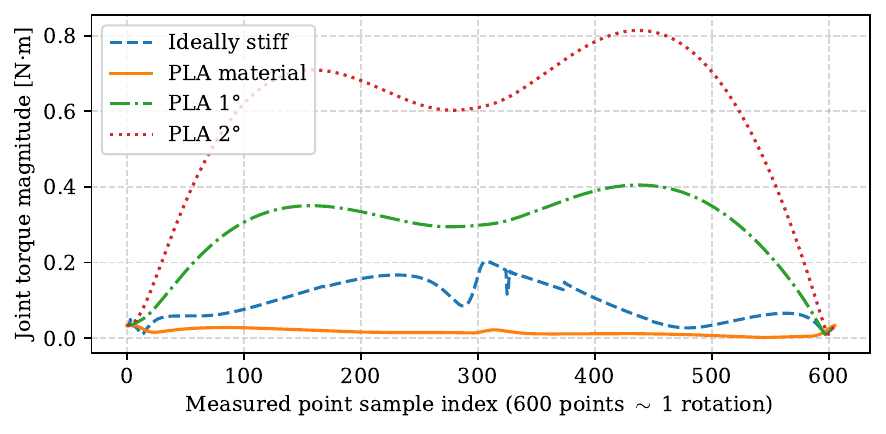}
        \caption{Joint torque magnitude at joint~3.}
        \label{fig:torque_joint3}
    \end{subfigure}
    \caption{Comparison of the deviation error from the simulated trajectories to the ideal trajectory (ideal -- zero deviation). The stiff material deviation is close to zero (all deviation values are bellow $0.001\,\mathrm{mm}$). The joint torque magnitude increases with misalignment angle, reflecting the constraint forces required to close the kinematic loop.}
    \label{fig:deviation_errors}
\end{figure*}

To validate the proposed flexible multibody modeling approach, the simulated trajectories of the end-effector were compared to the ideal trajectory obtained from the motion polynomial, as well as to the measured trajectories of the physical prototypes.

In Fig.~\ref{fig:compare_trajs} the comparison of the ideal and simulated trajectories is shown. The ideal trajectory is obtained from the motion polynomial obtained in Sect. \ref{sec:motion_recovery}, corresponding to the ideal DH parameters, while the simulated trajectory is obtained from the flexible multibody model in Exudyn. When the material properties in Exudyn are set to be ideally stiff (ideally rigid body), the simulated trajectory overlaps the ideal one, as expected. For the second simulation, the material properties are set to be of the PLA, introducing flexibility to the system, and the simulated trajectory is deviating from the ideal one, as it is expected in reality as well.

Fig.~\ref{fig:deviation_errors} shows the trajectory deviation from the theoretical motion curve given by Eq.~\eqref{eq:motion}. The maximum trajectory errors are bellow the following values: $0.001\,\mathrm{mm}$ in the ideally stiff (rigid) variant, $0.211\,\mathrm{mm}$ in the PLA variant, $1.215\,\mathrm{mm}$ in the PLA variant with $1^\circ$ misalignment, and $2.305\,\mathrm{mm}$ in the PLA variant with $2^\circ$ misalignment. The flexible PLA model with no misalignment yields a small error of $0.2\,\mathrm{mm}$, and the deviation grows consistently with increasing misalignment angle, confirming the expected trend. 
In~Fig.~\ref{fig:torque_joint3}, the torques of the third joint (driving joint) are plotted for reference. The initial and ending positions are the same as in~Fig.~\ref{fig:exu_setup}, with coupler link going first downwards. Therefore, the torque magnitude is growing slower during the first quarter of the motion as gravity acts in the same direction. The ideally rigid model requires higher driving torques than the flexible PLA model: since the overconstrained linkage enforces all redundant constraints exactly, larger internal constraint forces build up, which the driving joint must overcome. In the flexible model, link compliance partially absorbs these forces through elastic deformation.

In Tab.~\ref{tab:dh_deviation}, the summary of the calculated DH parameters using the methodology introduced in Sect.~\ref{sec:identification} is shown. The simulated results tend to show better accuracy, which is expected, as the joints are considered ideal and there is no 3D-printer inaccuracy or material shrinkage involved~\cite{Akba_2019dimensi}. 
The error of $d_i$ parameters, which should be by the Bennett condition zero, deviates from the ideal value the most; however, note that $d_i$ is the distance along joint axis $i$ of to the common perpendiculars with the adjacent axes, so even a small misalignment can cause a large distance offset. We therefore suggest that readers focus on the comparison of $a_i$ and $\alpha_i$ values. 

\definecolor{lvl1}{RGB}{255,255,255}
\definecolor{lvl2}{RGB}{255,250,210}
\definecolor{lvl3}{RGB}{255,230,160}
\definecolor{lvl4}{RGB}{255,190,150}
\definecolor{lvl5}{RGB}{255,160,140}
\newcommand{\dhd}[2]{%
  \pgfmathparse{abs(#2)}%
  \ifdim\pgfmathresult pt<0.3pt \cellcolor{lvl1}%
  \else\ifdim\pgfmathresult pt<1.0pt \cellcolor{lvl2}%
  \else\ifdim\pgfmathresult pt<2.0pt \cellcolor{lvl3}%
  \else\ifdim\pgfmathresult pt<6.0pt \cellcolor{lvl4}%
  \else \cellcolor{lvl5}%
  \fi\fi\fi\fi $#1$}
\newcommand{\dha}[2]{%
  \pgfmathparse{abs(#2)}%
  \ifdim\pgfmathresult pt<0.05pt \cellcolor{lvl1}%
  \else\ifdim\pgfmathresult pt<0.20pt \cellcolor{lvl2}%
  \else\ifdim\pgfmathresult pt<0.75pt \cellcolor{lvl3}%
  \else\ifdim\pgfmathresult pt<1.50pt \cellcolor{lvl4}%
  \else \cellcolor{lvl5}%
  \fi\fi\fi\fi $#1$}
\newcommand{\dhalpha}[2]{%
  \pgfmathparse{abs(#2)}%
  \ifdim\pgfmathresult pt<0.05pt \cellcolor{lvl1}%
  \else\ifdim\pgfmathresult pt<0.20pt \cellcolor{lvl2}%
  \else\ifdim\pgfmathresult pt<0.50pt \cellcolor{lvl3}%
  \else\ifdim\pgfmathresult pt<1.00pt \cellcolor{lvl4}%
  \else \cellcolor{lvl5}%
  \fi\fi\fi\fi $#1$}
\newcommand{\sig}[1]{\textcolor{gray}{$#1$}}
\begin{table}
\centering
\scriptsize
\caption{Identified DH parameters for all configurations. Lab results reported as mean~$\pm$~std~($\sigma$) from five measurements; simulations have no measurement uncertainty (the results are deterministic). Intentionally misaligned $\alpha_3$ values are bold. Cell color encodes the absolute deviation from the nominal value: white (negligible), light yellow (small), yellow (moderate), orange (large), red (very large). Thresholds are parameter-specific: for $a_i,\,d_i$ [mm], boundaries are set at $0.05/0.20/0.75/1.50$~mm ($a_i$) and $0.3/1.0/2.0/6.0$~mm ($d_i$); for $\alpha_i$ [deg], boundaries are set at $0.05/0.20/0.50/1.00$~deg.}
\label{tab:dh_deviation}
\setlength{\tabcolsep}{3pt}
\begin{tabular}{l c @{\hspace{6pt}} rr @{\hspace{6pt}} rr @{\hspace{6pt}} rr}
\toprule
\multirow{2}{*}{} & \multirow{2}{*}{$i$}
  & \multicolumn{2}{c}{$a_i$ [mm]}
  & \multicolumn{2}{c}{$\alpha_i$ [deg]}
  & \multicolumn{2}{c}{$d_i$ [mm]} \\
\cmidrule(lr){3-4}\cmidrule(lr){5-6}\cmidrule(lr){7-8}
  & & val & $\pm\,\sigma$ & val & $\pm\,\sigma$ & val & $\pm\,\sigma$ \\
\midrule
\multicolumn{8}{l}{\textit{Simulation Exudyn (nominal reference)}} \\
\midrule
\multirow{4}{*}{Ideally stiff}
  & 0 \; & \dha{110.000}{0.000} & & \dhalpha{150.000}{0.000} & & \dhd{0.000}{0.000}  & \\
  & 1 \; & \dha{220.000}{0.000} & & \dhalpha{90.000}{0.000}  & & \dhd{0.000}{0.000}  & \\
  & 2 \; & \dha{110.000}{0.000} & & \dhalpha{150.000}{0.000} & & \dhd{0.000}{0.000}  & \\
  & 3 \; & \dha{220.000}{0.000} & & \dhalpha{90.000}{0.000}  & & \dhd{0.000}{0.000}  & \\
\midrule
\multicolumn{8}{l}{\textit{Simulation Exudyn (PLA stiffness)}} \\
\midrule
\multirow{4}{*}{PLA}
  & 0 \; & \dha{110.004}{0.004}  & & \dhalpha{149.995}{0.005}  & & \dhd{0.251}{0.251}   & \\
  & 1 \; & \dha{219.989}{0.011}  & & \dhalpha{90.004}{0.004}   & & \dhd{-0.130}{0.130}  & \\
  & 2 \; & \dha{109.997}{0.003}  & & \dhalpha{149.989}{0.011}  & & \dhd{0.323}{0.323}   & \\
  & 3 \; & \dha{219.999}{0.001}  & & \dhalpha{90.010}{0.010}   & & \dhd{-0.006}{0.006}  & \\
\cmidrule(lr){3-7}
\multirow{4}{*}{PLA 1$^\circ$}
  & 0 \; & \dha{110.002}{0.002}  & & \dhalpha{149.839}{0.161}  & & \dhd{0.688}{0.688}   & \\
  & 1 \; & \dha{220.068}{0.068}  & & \dhalpha{90.680}{0.680}   & & \dhd{1.736}{1.736}  & \\
  & 2 \; & \dha{110.031}{0.031}  & & \dhalpha{149.942}{0.058}  & & \dhd{0.473}{0.473}   & \\
  & 3 \; & \dha{220.041}{0.041}  & & \dhalpha{\textbf{90.577}}{0.577}   & & \dhd{0.296}{0.296}  & \\
\cmidrule(lr){3-7}
\multirow{4}{*}{PLA 2$^\circ$}
  & 0 \; & \dha{109.999}{0.001}  & & \dhalpha{149.685}{0.315}  & & \dhd{1.628}{1.628}   & \\
  & 1 \; & \dha{220.136}{0.136}  & & \dhalpha{91.350}{1.350}   & & \dhd{3.370}{3.370}  & \\
  & 2 \; & \dha{110.061}{0.061}  & & \dhalpha{149.897}{0.103}  & & \dhd{0.601}{0.601}   & \\
  & 3 \; & \dha{220.081}{0.081}  & & \dhalpha{\textbf{91.140}}{1.140}   & & \dhd{0.586}{0.586}  & \\
\midrule
\multicolumn{8}{l}{\textit{Lab measurements (PLA prototypes)}} \\
\midrule
\multirow{4}{*}{PLA}
  & 0 \; & \dha{110.072}{0.072}  & \sig{\pm0.068} & \dhalpha{149.875}{0.125} & \sig{\pm0.053} & \dhd{-0.255}{0.255} & \sig{\pm0.182} \\
  & 1 \; & \dha{219.581}{0.419}  & \sig{\pm0.228} & \dhalpha{90.213}{0.213}  & \sig{\pm0.116} & \dhd{0.079}{0.079}  & \sig{\pm0.466} \\
  & 2 \; & \dha{109.956}{0.044}  & \sig{\pm0.043} & \dhalpha{149.909}{0.091} & \sig{\pm0.023} & \dhd{0.731}{0.731}  & \sig{\pm0.362} \\
  & 3 \; & \dha{220.165}{0.165}  & \sig{\pm0.035} & \dhalpha{90.135}{0.135}  & \sig{\pm0.194} & \dhd{0.497}{0.497}  & \sig{\pm0.254} \\
\cmidrule(lr){3-8}
\multirow{4}{*}{PLA 1$^\circ$}
  & 0 \; & \dha{109.926}{0.074}  & \sig{\pm0.180} & \dhalpha{150.165}{0.165} & \sig{\pm0.066} & \dhd{2.294}{2.294}  & \sig{\pm0.652} \\
  & 1 \; & \dha{220.756}{0.756}  & \sig{\pm0.234} & \dhalpha{89.976}{0.024}  & \sig{\pm0.087} & \dhd{1.756}{1.756}  & \sig{\pm0.777} \\
  & 2 \; & \dha{110.318}{0.318}  & \sig{\pm0.050} & \dhalpha{150.007}{0.007} & \sig{\pm0.011} & \dhd{0.808}{0.808}  & \sig{\pm0.261} \\
  & 3 \; & \dha{220.471}{0.471}  & \sig{\pm0.055} & \dhalpha{\textbf{90.438}}{0.438}  & \sig{\pm0.082} & \dhd{0.489}{0.489}  & \sig{\pm0.163} \\
\cmidrule(lr){3-8}
\multirow{4}{*}{PLA 2$^\circ$}
  & 0 \; & \dha{110.622}{0.622}  & \sig{\pm0.128} & \dhalpha{149.166}{0.834} & \sig{\pm0.058} & \dhd{8.137}{8.137}  & \sig{\pm0.985} \\
  & 1 \; & \dha{221.157}{1.157}  & \sig{\pm0.082} & \dhalpha{88.440}{1.560}  & \sig{\pm0.167} & \dhd{9.797}{9.797}  & \sig{\pm1.240} \\
  & 2 \; & \dha{111.501}{1.501}  & \sig{\pm0.108} & \dhalpha{149.239}{0.761} & \sig{\pm0.017} & \dhd{7.493}{7.493}  & \sig{\pm0.649} \\
  & 3 \; & \dha{220.547}{0.547}  & \sig{\pm0.035} & \dhalpha{\textbf{90.915}}{0.915}  & \sig{\pm0.087} & \dhd{5.039}{5.039}  & \sig{\pm0.781} \\
\midrule
\multicolumn{8}{l}{\textit{Lab measurements (other prototypes)}} \\
\midrule
\multirow{4}{*}{Cardboard}
  & 0 \; & \dha{110.758}{0.758}  & \sig{\pm0.055} & \dhalpha{149.197}{0.803} & \sig{\pm0.074} & \dhd{1.328}{1.328}  & \sig{\pm0.240} \\
  & 1 \; & \dha{218.585}{1.415}  & \sig{\pm0.174} & \dhalpha{91.009}{1.009}  & \sig{\pm0.056} & \dhd{2.167}{2.167}  & \sig{\pm0.467} \\
  & 2 \; & \dha{111.419}{1.419}  & \sig{\pm0.011} & \dhalpha{149.634}{0.366} & \sig{\pm0.011} & \dhd{-0.848}{0.848} & \sig{\pm0.125} \\
  & 3 \; & \dha{219.097}{0.903}  & \sig{\pm0.006} & \dhalpha{90.877}{0.877}  & \sig{\pm0.076} & \dhd{-0.924}{0.924} & \sig{\pm0.105} \\
\cmidrule(lr){3-8}
\multirow{4}{*}{Bamboo}
  & 0 \; & \dha{109.423}{0.577}  & \sig{\pm0.254} & \dhalpha{150.718}{0.718} & \sig{\pm0.089} & \dhd{0.363}{0.363}  & \sig{\pm1.518} \\
  & 1 \; & \dha{222.262}{2.262}  & \sig{\pm0.343} & \dhalpha{89.665}{0.335}  & \sig{\pm0.244} & \dhd{-0.112}{0.112} & \sig{\pm1.140} \\
  & 2 \; & \dha{109.007}{0.993}  & \sig{\pm0.171} & \dhalpha{150.435}{0.435} & \sig{\pm0.025} & \dhd{-0.110}{0.110} & \sig{\pm0.438} \\
  & 3 \; & \dha{221.764}{1.764}  & \sig{\pm0.079} & \dhalpha{89.822}{0.178}  & \sig{\pm0.088} & \dhd{-0.247}{0.247} & \sig{\pm0.286} \\
\bottomrule
\end{tabular}
\end{table}

\begin{figure}[pos=b]
    \centering
    \includegraphics[width=\linewidth]{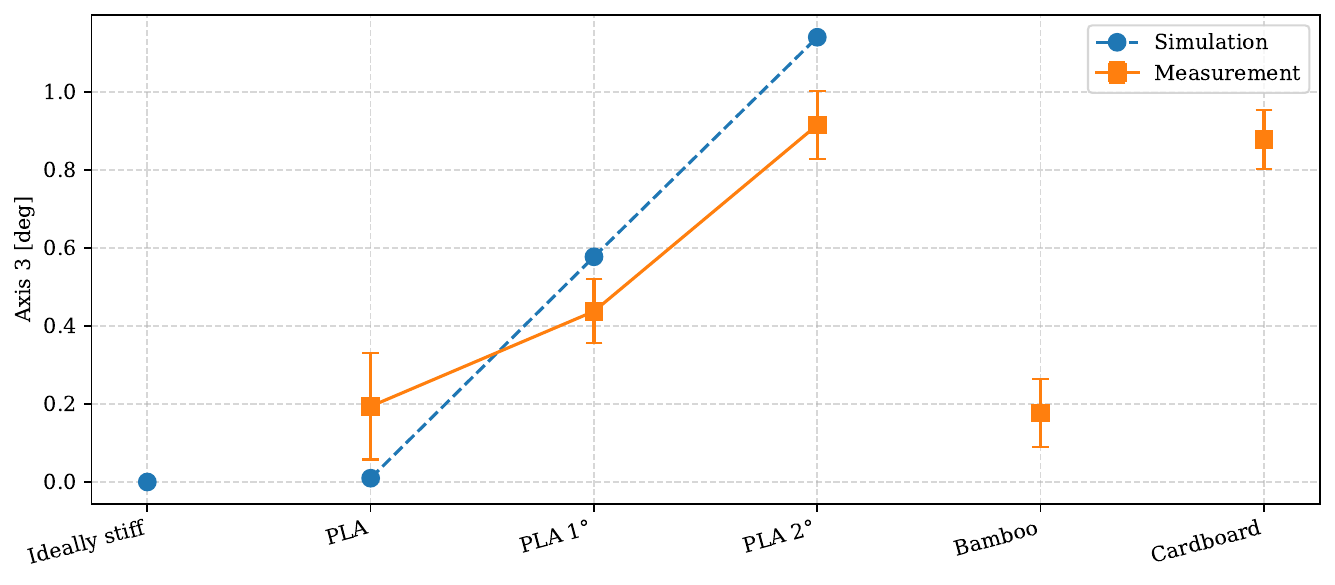}
    \caption{Angular deviation of joint axis~3 identified from simulated (circles, dashed) and measured (squares, solid) coupler trajectories for all prototype configurations. Each measurement marker shows the mean over five repeated runs; error bars indicate~$\pm$1 standard deviation. The simulation reproduces the trend of increasing deviation with growing joint-axis misalignment.}
    \label{fig:results_axis3}
\end{figure}

In both simulation and experiment we can see similar trend of rising error in the case of the intentional misalignment. In particular, 
comparison of the simulated and measured axis deviations across all prototype configurations is summarized in Fig.~\ref{fig:results_axis3}, which shows the angular deviation of joint axis~3 identified from the link trajectories. Each measurement point represents the mean over five repeated runs, with error bars indicating~$\pm$1 standard deviation. The simulation reproduces the trend observed in the physical measurements: the nominal prototypes exhibit small deviations, while the intentionally misaligned configurations show increasing angular error proportional to the introduced misalignment. The bamboo and paper prototypes are included as measurement-only reference cases, as no material-specific simulation was performed for these configurations.

\subsection{Supplementary material}\label{sec:supplementary_mat} 

Supplementary material accompanying this work is provided at~\cite{suppl_mat}, including all data, simulation code, and analysis scripts required to reproduce the reported results. This includes raw and pre-processed VICON motion-capture measurements for four physical prototypes (bamboo, cardboard, and PLA links with and without intentional misalignment), the FEM-based flexible-body simulation model built with Exudyn, and the analysis and plotting scripts used to extract DH parameters, compute axis-3 elevation, and generate the trajectory deviation, joint torque, and Bennett condition comparisons presented in the paper. A detailed README describes the folder structure, data formats, and instructions for reproducing each the presented results.

\section{Discussion}\label{sec:discussion}

Fig.~\ref{fig:results_axis3} shows that deviations of measured prototypes are smaller than simulated misaligned variants, despite the larger DH parameter deviations reported in Tab.~\ref{tab:dh_deviation}. We believe this is a genuine effect: overconstrained mechanisms settle into a minimum-strain-energy configuration by exploiting the mechanism's flexibility (joint clearance, material compliance).

This becomes more intuitive when considering the stress introduced by any deviation from the ideal geometry -- in principle, the mechanisms should not be assemblable at all, and can only be forced together through the compliance of its parts.
Even if misaligned, the mechanism simply ``feels comfortable'' in the geometrically precise state, where it has an absolute freedom along the 1-DoF that it was designed for. The stress is therefore distributed in all assembly by exploiting the available flexibility. Tab.~\ref{tab:dh_deviation} demonstrates this clearly for the misaligned mechanisms -- although the  $\alpha_3$ value should be naturally equal 91° for PLA 1° and 92° for PLA 2°, respectively, the error propagated to other DH parameters instead, in a symmetric way. This can be also seen in~Tab.~\ref{tab:bennett_cond_comparison}, where the Bennett ratios are compared. In particular, for the Bamboo mechanism, the entire structure was deformed so notably that the Bennett ratio is $0.49$ instead of the ideal $0.5$, while still delivering the moveability close to the desired motion.

\definecolor{lvl1}{RGB}{255,255,255}
\definecolor{lvl2}{RGB}{255,250,210}
\definecolor{lvl3}{RGB}{255,230,160}
\definecolor{lvl4}{RGB}{255,190,150}
\newcommand{\bval}[2]{%
  \pgfmathparse{abs(#2)}%
  \ifdim\pgfmathresult pt<0.001pt \cellcolor{lvl1}%
  \else\ifdim\pgfmathresult pt<0.005pt \cellcolor{lvl2}%
  \else\ifdim\pgfmathresult pt<0.01pt \cellcolor{lvl3}%
  \else \cellcolor{lvl4}%
  \fi\fi\fi $#1$}
\begin{table}[h]
\centering
\footnotesize
\caption{Bennett condition ratios for all configurations. Cell shading indicates magnitude of deviation from the ideal value (white -- negligible ($<0.001$), light yellow -- small ($<0.005$), yellow -- moderate ($<0.01$), orange -- large ($\geq0.01$) deviation).}
\label{tab:bennett_cond_comparison}
\setlength{\tabcolsep}{4pt}
\begin{tabular}{l @{\hspace{8pt}} cc @{\hspace{8pt}} cc @{\hspace{4pt}} c}
\toprule
  & \multicolumn{2}{c}{Link length ratios}
  & \multicolumn{2}{c}{Twist angle ratios}
  & \\
\cmidrule(lr){2-3}\cmidrule(lr){4-5}
Mechanism
  & $a_0/a_1$ & $a_2/a_3$
  & $\sin\alpha_0/\sin\alpha_1$ & $\sin\alpha_2/\sin\alpha_3$
  \\
\midrule
\multicolumn{6}{l}{\textit{Simulation Exudyn (nominal reference)}} \\
\midrule
Ideally stiff
  & \bval{0.5000}{0.0000} & \bval{0.5000}{0.0000}
  & \bval{0.5000}{0.0000} & \bval{0.5000}{0.0000}
  \\
\midrule
\multicolumn{6}{l}{\textit{Simulation Exudyn (PLA stiffness)}} \\
\midrule
PLA
  & \bval{0.5001}{0.0001} & \bval{0.5000}{0.0000}
  & \bval{0.5001}{0.0001} & \bval{0.5002}{0.0002}
  \\
PLA 1$^\circ$
  & \bval{0.4999}{0.0001} & \bval{0.5000}{0.0000}
  & \bval{0.5025}{0.0025} & \bval{0.5009}{0.0009}
  \\
PLA 2$^\circ$
  & \bval{0.4997}{0.0003} & \bval{0.5001}{0.0001}
  & \bval{0.5049}{0.0049} & \bval{0.5017}{0.0017}
  \\
\midrule
\multicolumn{6}{l}{\textit{Lab measurements (PLA prototypes)}} \\
\midrule
PLA
  & \bval{0.5013}{0.0013} & \bval{0.4994}{0.0006}
  & \bval{0.5019}{0.0019} & \bval{0.5014}{0.0014}
  \\
PLA 1$^\circ$
  & \bval{0.4980}{0.0020} & \bval{0.5004}{0.0004}
  & \bval{0.4975}{0.0025} & \bval{0.4999}{0.0001}
  \\
PLA 2$^\circ$
  & \bval{0.5002}{0.0002} & \bval{0.5056}{0.0056}
  & \bval{0.5127}{0.0127} & \bval{0.5115}{0.0115}
  \\
\midrule
\multicolumn{6}{l}{\textit{Lab measurements (other prototypes)}} \\
\midrule
Cardboard
  & \bval{0.5067}{0.0067} & \bval{0.5085}{0.0085}
  & \bval{0.5122}{0.0122} & \bval{0.5056}{0.0056}
  \\
Bamboo
  & \bval{0.4923}{0.0077} & \bval{0.4915}{0.0085}
  & \bval{0.4891}{0.0109} & \bval{0.4934}{0.0066}
  \\
\bottomrule
\end{tabular}
\end{table}

Despite the demonstrated ability of the mechanism to compensate for geometric inaccuracies through structural deformation, it is important to emphasize that this capability comes at the cost of reduced motion accuracy. In other words, while the mechanism remains operational even in the presence of deviations from the ideal geometry, the end-effector trajectory systematically deviates from the nominal solution due to elastic deformations. This effect is particularly evident when comparing ideal, simulated and experimentally measured trajectories, where the deviation increases with the level of geometric inaccuracy and structural compliance.

From this perspective, it becomes clear that, for high-precision applications, these mechanisms remain viable, but only if they are manufactured with sufficient precision. The self-assembling tendency that makes them tolerant of coarse assembly is not a substitute for geometric accuracy when tight positioning is required. Therefore, the design of such systems must account not only for kinematic synthesis but also for the mechanical properties of the structure, particularly its stiffness and sensitivity to loading conditions. Our flexible-body modeling approach can address exactly this.

Future research should focus on the development of models capable of predicting the deformation of the end-effector trajectory as a function of structural and material parameters. Such models could serve as a foundation for the design of mechanisms with stiffness optimized for critical configurations, or for the development of compensation strategies aimed at reducing the influence of deformation on motion accuracy.

Last but not least, we focused on the 4R Bennett mechanisms although the presented methodology is applicable on overconstrained single-loop 6R mechanisms~\cite{Hegeds2015}, too. However, from the experience with 3D-printed prototypes as in~\cite[Fig.~1]{huczala2026-ark}, a crucial problem was observed -- these mechanisms tend to suffer from singularities and snap to isolated, theoretically rigid, configurations, which are ``nearby'' their motion curve. 
The modeling problem is therefore even more challenging, and the flexible multibody method with a specific monitoring of mechanism buckling seems to be an appropriate tool to approach this problematic behavior and analyze if a particular 6R mechanism will suffer from the unintended configurations. Nevertheless, this requires much more work and will be investigated by the authors in the future.

\section{Conclusion} 

This study presented a flexible multibody modeling framework for overconstrained rational single-loop linkages, addressing a gap in the current state of the art where rigid body approaches fail to handle redundant constraints in a physically meaningful way. The proposed methodology was validated against 3D-printed PLA prototypes, including variants with intentional joint-axis misalignment.

The flexible modeling approach demonstrated clear advantages over rigid body methods: by accounting for link compliance, the framework captures the physical behavior of overconstrained mechanisms. The model provides access to dynamic parameters (joint torques and constraint forces, etc.) that are essential for optimal control, and allows simulation parameters to be tailored to a specific design scenario prior to manufacturing, reducing the risk of costly iterations.

A key finding of this study is that overconstrained mechanisms exhibit a self-assembling tendency: structural compliance allows the mechanism to settle into the geometrically ideal configuration even in the presence of assembly inaccuracies. Two supplementary contributions were introduced: a method for recovering rational motion polynomials from DH parameters, and a novel kinematic parameter identification methodology based on motion capture, PCA, and circle fitting. The combination of the methods provided can serve as an accuracy benchmark specifically adapted to single-loop linkage architectures.

Future work will extend the framework to overconstrained six-bar mechanisms, where singularity and buckling behavior pose additional modeling challenges. Further directions include design parameter optimization using the flexible model, compensation strategies for trajectory deviations caused by link compliance, and advanced singularity analysis that accounts for physical material parameters.

\section*{Supplementary Material}

For more details, see Sect.~\ref{sec:supplementary_mat}. The material is available in~\cite{suppl_mat} D. Huczala, M. Pieber, J. Gerstmayr, A. Mair, F. Schulte, S. Glass, T. Poštulka, A. Vysocký, M. Pfurner, Flexible-body linkage modeling: Supplementary material (2026), \doi{10.5281/zenodo.22076383}.

\section*{Acknowledgments}

This work was supported by the InnoCORE program of the Ministry of Science and ICT (1.260018.01).

This work was supported by the European Regional Development Fund under the project Research Platform for Digital Transformation and Society 5.0 CZ.02.01.01/00/ 23\_021/0012599 within the Jan Amos Komensky Operational Program.

We gratefully acknowledge the support of the bilateral Austrian–Korean project “Surface generation via integrable evolution of curves,” which is funded by OeAD-GmbH (Austria’s Agency for Education and Internationalization, project no. KR 02/2025), and the National Research Foundation of Korea (project no. RS-2025-1435299).







\printcredits

\section*{Declaration of Generative AI and AI-Assisted Technologies in the Writing Process}

During the preparation of this work, the authors used Claude and ChatGPT tools in order to improve the readability and clarity of the manuscript. After using this tool, the authors reviewed and edited the content as needed and take full responsibility for the content of the published article.

\bibliographystyle{elsarticle-num}

\bibliography{references}



\end{document}